\documentclass{article}

\usepackage{PRIMEarxiv}

\usepackage[utf8]{inputenc} 
\usepackage[T1]{fontenc}    
\usepackage{hyperref}       
\usepackage{url}            
\usepackage{booktabs}       
\usepackage{amsfonts}       
\usepackage{nicefrac}       
\usepackage{microtype}      
\usepackage{lipsum}
\usepackage{fancyhdr}       
\usepackage{graphicx}       
\graphicspath{{figs/}}      

\usepackage{amsmath, amssymb, amsthm}

\usepackage{longtable}
\usepackage{pdflscape}
\usepackage{natbib}
\usepackage{colortbl, xcolor}

\usepackage{booktabs, tabularx, array}
\usepackage{pgfmath}
\definecolor{Teal}{HTML}{00C1D5}
\definecolor{Purple}{HTML}{AA4AC4}
\definecolor{startcol}{HTML}{56B4E9}
\definecolor{endcol}{HTML}{E69F00} 

\newcolumntype{R}{>{\raggedleft\arraybackslash}X}

\newcommand{\shadefromto}[3]{%
  \pgfmathsetmacro{\pct}{100*(#1-#2)/max(1e-9,#3-#2)}%
  \edef\mix{\noexpand\cellcolor{startcol!\pct!endcol!50}}%
  \mix #1%
}

\newcommand{\shadeabserr}[2]{%
  \pgfmathsetmacro{\pct}{100*abs(#1)/max(1e-9,#2)}%
  \edef\mix{\noexpand\cellcolor{startcol!\pct!endcol!50}}%
  \mix #1%
}

\title{Do We Really Even Need Data?\\ A Modern Look at Drawing Inference with Predicted Data}

\author{
  Stephen Salerno \\
  Public Health Sciences Division \\
  Fred Hutchinson Cancer Center \\
  Seattle, WA \\
  \texttt{ssalerno@fredhutch.org} \\
  \And
  Kentaro Hoffman \\
  Department of Statistics \\
  University of Washington \\
  Seattle, WA \\
  \texttt{khoffm3@uw.edu} \\
  \And
  Awan Afiaz \\
  Department of Biostatistics \\
  University of Washington \\
  Public Health Sciences Division \\
  Fred Hutchinson Cancer Center \\
  Seattle, WA \\
  \texttt{aafiaz@uw.edu} \\
  \AND
  Anna Neufeld \\
  Department of Mathematics and Statistics \\
  Williams College \\
  Williamstown, MA
  \And
  Tyler H.~McCormick$^\dag$\\
  Department of Statistics \\
  Department of Sociology \\
  University of Washington \\
  Seattle, WA \\
  \texttt{tylermc@uw.edu} \\
  \And
  Jeffrey T.~Leek$^\dag$ \\
  Public Health Sciences Division \\
  Fred Hutchinson Cancer Center \\
  Seattle, WA \\
  \texttt{jtleek@fredhutch.org} \\
}

\begin{document}

\maketitle


\begin{abstract}
    As artificial intelligence and machine learning tools become more accessible, and scientists face new obstacles to data collection (e.g.,~rising costs, declining survey response rates), researchers increasingly use predictions from pre-trained algorithms as substitutes for missing or unobserved data. Though appealing for financial and logistical reasons, using standard tools for inference can misrepresent the association between independent variables and the outcome of interest when the true, unobserved outcome is replaced by a predicted value. In this paper, we characterize the statistical challenges inherent to drawing {\it inference with predicted data} (IPD) and show that high predictive accuracy does not guarantee valid downstream inference. We show that all such failures reduce to statistical notions of (i) {\it bias}, when predictions systematically shift the estimand or distort relationships among variables, and (ii) {\it variance}, when uncertainty from the prediction model and the intrinsic variability of the true data are ignored. We then review recent methods for conducting IPD and discuss how this framework is deeply rooted in classical statistical theory. We then comment on some open questions and interesting avenues for future work in this area, and end with some comments on how to use predicted data in scientific studies that is both transparent and statistically principled.
\end{abstract}


\section*{Media Summary}

Artificial intelligence and machine learning are increasingly used to ``fill in'' missing or expensive data, by predicting outcomes, labels, or covariates we cannot directly observe. Those predictions can be accurate for individuals, yet still mislead scientific conclusions if we treat them as if they were real measurements. This review explains why predictions can carry bias into analyses, make uncertainty look smaller than it is, and subtly distort relationships between variables. We then show how a new class of assumption-lean methods seek to address these issues by combining a small set of gold-standard labels with potentially many imperfect predictions.


\section{If you had never seen a rhinoceros, how would you draw one?}
\label{sec:1}

\begin{figure}[!ht]
    \centering
    \includegraphics[height = 2in]{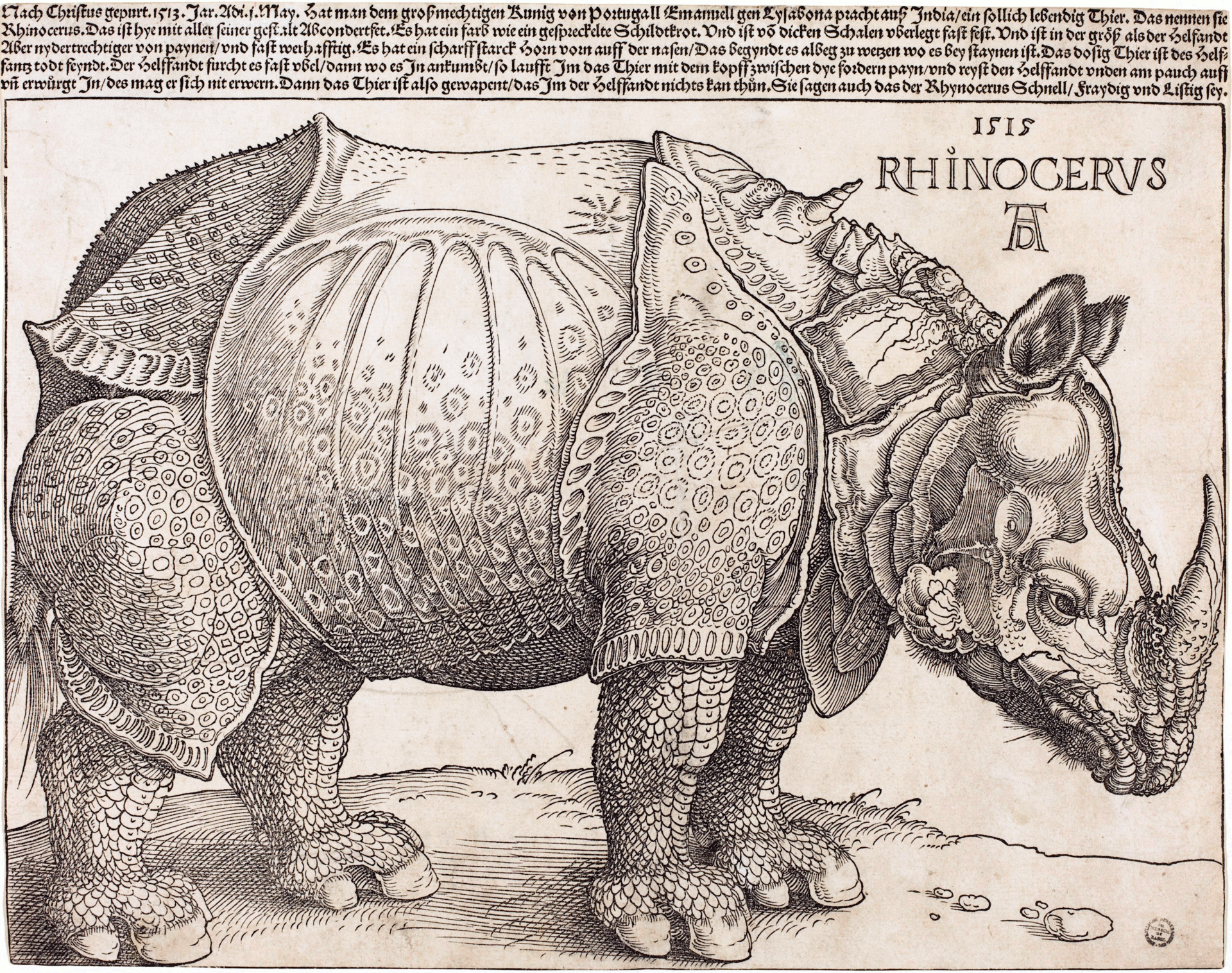}
    \includegraphics[height = 2in]{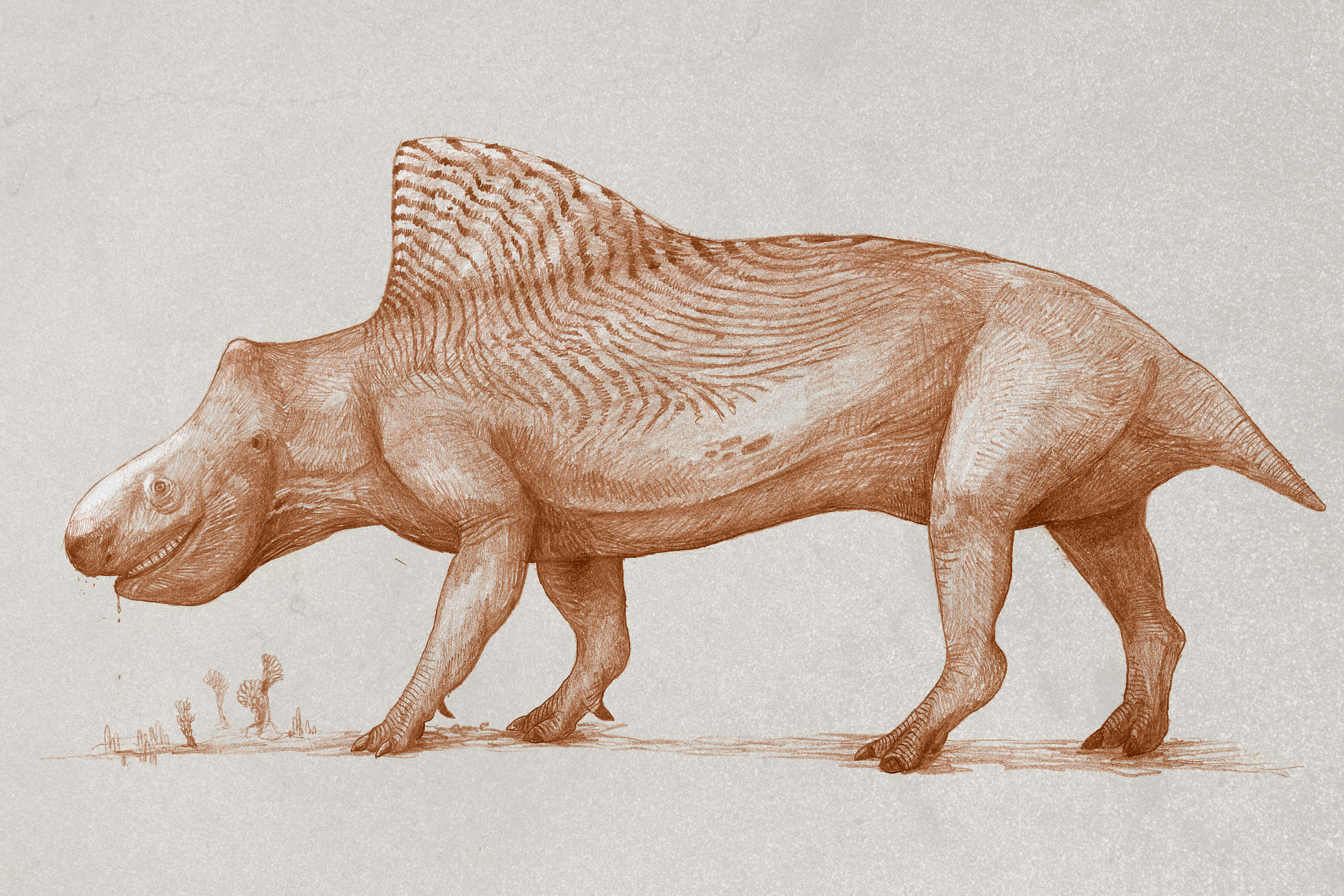}
    \caption{Artist renderings of a rhinoceros based on limited information. Left: Albrecht D{\"u}rer's {\it The Rhinoceros}, woodcut (1515); Right: C.M.~K{\"o}semen's paleoart reconstruction of a rhinoceros based on its skeleton (2012).}
    \label{fig:rhino}
\end{figure}

Albrecht D{\"u}rer had never seen a rhino in 1515 when he made his woodcutting (Fig.~\ref{fig:rhino}, left panel). Instead, he used statistical {\it inference}. He gathered a sample of descriptions from several people who had seen a rhino to create a unified summary. While not entirely accurate, D{\"u}rer's work remained the prevailing European understanding of a rhino through the 18th century \cite{rookmaaker2005review}. 

C.M.~K{\"o}semen had seen a rhino before creating his 2012 depiction (Fig.~\ref{fig:rhino}, right panel).  His goal, though, was to draw the contemporary animal in the style of common illustrations of prehistoric animals.  Since no living human has ever seen these prehistoric creatures, there's no sample of direct accounts to be had.  Instead, we draw dinosaurs and similar creatures based on fossilized remains, which capture some of the key features that characterize appearance but omit certain fundamental details~\citep{conway2012all}. K{\"o}semen's rhino shows us what an animal salient to our current consciousness, a rhinoceros, would look like if we applied the {\it prediction} model we used for dinosaurs, i.e., generating an unseen outcome based on potentially limited information. Like K{\"o}semen's rhino, artificial intelligence and machine learning (AI/ML) can generate data based on potentially limited information, raising critical questions about accuracy and bias when these predictions are then used for downstream statistical {\it inference}.

\subsection{Sightings of K{\"o}semen's Rhino}

The population of K{\"o}semen's rhinos is exploding across virtually all scientific settings. As AI/ML becomes embedded across scientific disciplines, the line between empirical and synthetic data has blurred. With rising data collection costs, declining survey response rates, slow annotation processes, and strict privacy regulations, researchers often turn to pretrained models to generate predicted values that augment or replace observed values. These models can generate complex outputs with high accuracy, and their widespread adoption has allowed researchers to seemingly `impute' hard-to-measure data across multiple domains.

Machine learning predictions increasingly support patient diagnosis, risk stratification, and treatment recommendations using electronic health record (EHR), imaging, and genetic data. For example, we can detect retinal disease and malignancies from imaging data~\citep{de2018clinically, mckinney2020international}. Computer vision enables automated karyotyping from metaphase spreads~\citep{shamsi2022karyotype, fang202342}. Natural language processing (NLP) can identify metastatic cancers from clinical notes~\citep{yang2022identification, alba2021ascertainment, yamoah2022racial}. Furthermore, AI/ML-enhanced pragmatic trials enable more efficient and externally valid study designs~\citep{gamerman2019pragmatic, williams2015pragmatic}. While these examples hold great promise for improving healthcare outcomes, biased predictions can lead to incorrect diagnoses or treatments and can widen health disparities, especially when training data underrepresent key groups~\citep{obermeyer2019dissecting}. 

Across broader `omics research, AI/ML methods are used to infer missing biological information, predict molecular interactions, and uncover complex regulatory mechanisms. We frequently predict missing phenotypic information based on partial genomic sequences~\citep{behravan2018machine, liu2017case, arumugam2011enterotypes, gamazon2015gene}. Transcriptome-wide association studies link imputed gene expression levels to disease outcomes~\citep{gusev2019transcriptome}. In epigenomics, we can identify regulatory elements and chromatin accessibility patterns from sequencing data~\citep{avsec2021effective}. AI-driven models can also predict protein structures, interactions, and functions in proteomics~\citep{jumper2021highly} and discover new disease biomarkers and metabolic pathways~\citep{qiu2023small, barberis2022precision, chi2024artificial}. While these approaches facilitate novel drug discovery and personalized medicine, they create statistical problems that lead to more false positive findings and difficulties when interpreting the biological meaning of those findings.

Beyond biomedicine, environmental models uses historic satellite imagery, ground sensor readings, and climate records to generate forecasts that guide early-warning systems, targeted resource allocation, and disaster preparedness strategies for hurricanes, droughts, and wildfires~\citep{rasp2024weatherbench, rolnick2022tackling}. However, even small biases in these models can lead to problematic downstream decision-making, which can have considerable effects on policy decisions and resource planning~\citep{schneider2023harnessing}. Demographers use AI/ML in low-resource settings to estimate causes of death and other vital statistics from verbal autopsy surveys~\citep{murray2014using, chen2025lava} or satellite-derived features such as nighttime light intensity~\citep{evans2013estimates}. These approaches provide critical public health information in settings where civil registration systems are incomplete or non-existent, but they also complicate statistical inference, especially when their outputs are used in downstream hypotheses-driven analyses~\citep{fan2024narratives}.

Social scientists use AI-driven sentiment analysis of social media content to infer public opinion, track social movements, and analyze discourse trends in real time~\citep{barbera2015tweeting, grimmer2013text}. Modern methods are now being applied to historical texts and digitized archival materials as well~\citep{michel2011quantitative}. Further still, census data and street-level photography are increasingly being used to predict an individual's race based on their neighborhood composition~\citep{gebru2017using}. Economists use demand forecasting models to predict market trends based on satellite imagery of shipping container ports~\citep{yu2023eye}. Central banks use `nowcasting' models that incorporate high-frequency indicators, such as web-scraped prices and financial sentiment scores, to guide monetary policy~\citep{banbura2013now}. However, these models rely on assumptions about consumer behavior and macroeconomic stability, and prediction errors can lead to misinformed economic policy~\citep{hoffman2024some}.

In fact, Table \ref{tab:examples} gives fifty examples of scientific inquiry where AI/ML-generated data can enter downstream analyses. A common thread is that synthetic data are used alongside empirical data, without distinction, across many disciplines. While these applications demonstrate the remarkable utility of AI/ML, they also present a statistical challenge:~{\it when AI/ML-generated data are used in subsequent analyses, how do we ensure valid downstream inference?} 

While our title is a bit tongue-in-cheek, it highlights two points:~(1) increasing reliance on AI/ML-generated data in scientific research raises important questions about inferential validity, and (2) drawing {\it inference with predicted data} (IPD) is a rapidly evolving area, driven by the need for rigorous methods. Our goal is to clarify the statistical foundations for IPD. We review this emerging area and explore how recent methods correct for bias and/or additional uncertainty. We also contextualize this framework within classical fields such as survey sampling, missing data, measurement error, and semi-supervised learning, emphasizing how study design continues to play a critical role in modern inference. We will work through case studies on population-level inference in public health and politics to demonstrate the practical implications of IPD. Lastly, we conclude with some considerations on the use of AI/ML to generate synthetic data and outline our thoughts on the future of this field, given that ``K{\"o}semen's rhino'' now inhabits every corner of modern science.


\section{The Setup and The Problem}
\label{sec:2}

\subsection{The Setup} 
\begin{figure}[!ht]
    \centering
    \includegraphics[width=\linewidth]{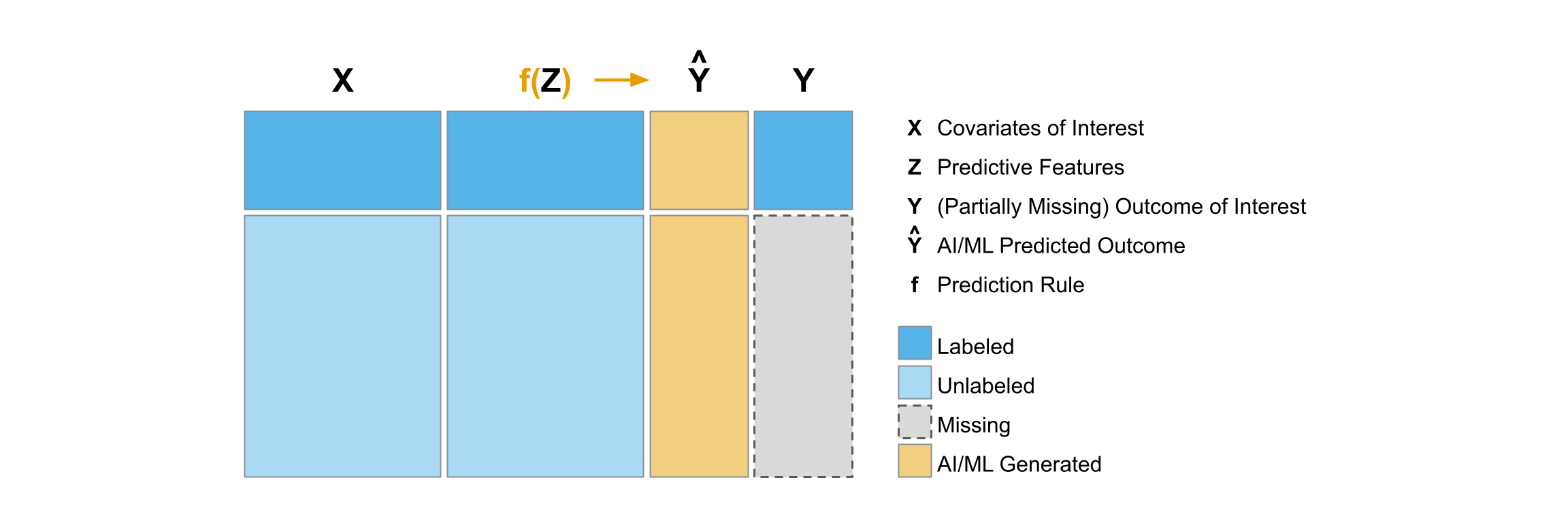}
    \caption{Overview of the setup for inference with predicted data.}
    \label{fig:data}
\end{figure}

Consider a study with $n$ observations of $(Y, \boldsymbol{X}, \boldsymbol{Z})$, where our outcome, $Y,$ is partially missing due to constraints such as time or cost, $\boldsymbol{X}$ are covariates for downstream inference, $\boldsymbol{Z}$ are predictors used by a pretrained rule, $\hat{f}: \mathcal{Z} \mapsto \mathcal{Y}$, and $\hat{Y} = \hat{f}(\boldsymbol{Z})$ are predictions of $Y$. We make no assumptions on $\hat{f}$, only that it was trained independently of our sample~\citep[e.g.,][]{jumper2021highly, caruccio2024can}, which consists of a labeled, $\mathcal{L} = \{(Y_i, \hat{Y}_i, \boldsymbol{X}^\top_i, \boldsymbol{Z}^\top_i)\}_{i = 1}^{n_l}$, and an unlabeled, $\mathcal{U} = \{(\hat{Y}_i, \boldsymbol{X}^\top_i, \boldsymbol{Z}^\top_i)\}_{i = n_l + 1}^{n}$, subset, with $n = n_l + n_u$, and $n_l / n \to \rho \in (0,1)$ as $n \to \infty$ (Figure \ref{fig:data}). In practice, $\boldsymbol{X}$ may overlap $\boldsymbol{Z}$, but we do not require this. We focus on the case where $\boldsymbol{X}$ and $\boldsymbol{Z}$ are fully observed, though recent works explore more general missingness patterns. Further, we focus on settings where $\mathcal{L}$ and $\mathcal{U}$ are i.i.d., though recent works have relaxed this as well.

\begin{landscape}

\vspace*{-8ex}

\tiny

\begin{longtable}[t]{llp{4cm}p{4cm}p{4cm}p{4cm}}
\caption{Fifty examples of scientific inquiry where AI/ML-generated variables can enter downstream analyses.} \\
\label{tab:examples} \\
\toprule
{\bf Domain} & {\bf Citation} & {\bf Predicted Outcome} & {\bf Features} & {\bf Prediction Function} & {\bf Downstream Target} \\
\midrule
Demography    & \citet{chen2025lava}                   & Cause of Death                & Narrative Summaries from Caregivers               & LLMs                            & Univariate Associations \\
Demography    & \citet{evans2013estimates}             & PM 2.5 Exposure               & Satellite Images                                  & Chemical Transport Model        & Relative Risks \\
Demography    & \citet{fan2024narratives}              & Cause of Death                & Narrative Summaries from Caregivers               & LLMs, SVM, Random Forests       & Regression Coefficients \\
Demography    & \citet{murray2014using}                & Cause of Death                & Structured Verbal Autopsy Interview               & Random Forests, Bayesian Models & Population Proportions \\
Economics     & \citet{athey2025surrogate}             & Long-Term Treatment Effect    & Short-Term Employment, Earnings, Aid Receipt      & Linear Regression               & Double-Matching Estimator \\
Economics     & \citet{banbura2013now}                 & Quarterly GDP Growth          & Daily/Weekly Economic Indicators                  & Nowcastings Models              & Time Series Forecasts \\ 
Economics     & \citet{jean2016combining}              & Nighttime Light, Expenditures & Satellite Images                                  & CNNs                            & Ridge Regression Coefficients \\
Economics     & \citet{yu2023eye}                      & Number of Shipping Containers & Satellite Images of Ports                         & CNNs                            & Time Series Forecasts \\
Engineering   & \citet{jiang2020data}                  & Binary Machine Fault          & 22 Measurement Variables                          & Multi-GAN + DAC	             & Regression Coefficients \\
Environment   & \citet{rasp2024weatherbench}           & Weather Forecast              & Historic Weather Metrics (e.g., Precipitation)    & CNNs + Autoregressive Models    & Time Series Forecasts \\
Environment   & \citet{schneider2023harnessing}        & Climate Forecasts             & Observed and Simulated Climate/Earth Systems Data & Ensemble Climate Simulators     & Hazard Ratios \\
`Omics        & \citet{alipanahi2021large}             & Vertical Cup-to-Disc Ratio    & Color Fundus Photographs                          & Ensemble of CNNs                & GWAS, Regression Coefficients \\
`Omics        & \citet{an2023deep}                     & Missing Phenotypes	           & Phenotypes, Demographics, Labs, Imaging, Meds     & Autoencoder                     & GWAS \\
`Omics        & \citet{arumugam2011enterotypes}        & Microbiome Enterotypes        & Sequenced Metagenomes                             & Unsupervised Clustering         & Correlations, Fisher's Exact Test\\
`Omics        & \citet{avsec2021effective}             & Gene Expression               & DNA Sequences                                     & DNNs                            & GWAS \\
`Omics        & \citet{barberis2022precision}          & Disease Diagnosis, Prognosis  & Metabolomic Biomarkers                            & DNNs, SVMs, Random Forests      & Risk Stratification \\
`Omics        & \citet{behravan2018machine}            & Case-Control Status           & SNPs                                              & Gradient Boosting               & Gene-Interaction Mapping \\
`Omics        & \citet{chi2024artificial}              & Disease Diagnosis, Prognosis  & Metabolomic Biomarkers                            & DNNs, SVMs, Random Forests      & Risk Stratification, Regression Coefficents \\
`Omics        & \citet{ellis2018improving}             & Biological Phenotypes         & Expressed Regions from RNA-Seq Data               & Linear Models                   & Regression Coefficients \\
`Omics        & \citet{gamazon2015gene}                & Gene Expression               & SNPs                                              & Elastic Net                     & Regression Coefficients \\
`Omics        & \citet{gusev2019transcriptome}         & Gene Expression               & SNPs                                              & Penalized Regression            & Transcriptome-Wide Associations \\
`Omics        & \citet{jumper2021highly}               & 3D Protein Structures         & Amino Acid Sequences, 3D Atom Coordinates         & DNNs                            & Structural/Functional Inference \\	
`Omics        & \citet{kelley2018sequential}           & Epigenetic Profiles           & DNA Sequences	                                   & CNNs                            & Regression Coefficients \\
`Omics        & \citet{liu2017case}                    & Missing Genotypes             & Sequenced Genotypes                               & Genotype Imputation             & Regression Coefficients \\
`Omics        & \citet{zhou2018deep}                   & Disease Risk                  & DNA Sequences                                     & CNNs + Ridge Regression         & Regression Coefficients \\
Medicine      & \citet{alba2021ascertainment}          & Metastatic Status             & Clinician Notes                                   & NLP                             & Regression Coefficients \\
Medicine      & \citet{arvanitis2022method}            & Synthetic Patient Records     & Clinical Features, ICD-9 Codes	                   & GAN	                         & Regression Coefficients \\	
Medicine      & \citet{cosentino2023inference}         & COPD Liability Score          & Raw Flow-Volume Spirograms                        & CNNs                            & GWAS, Survival Estimates \\
Medicine      & \citet{dahl2023phenotype}              & Lifetime MDD Status           & Phenotypes, Comorbidities, Demographics, SES      & PCA + DNNs                      & GWAS \\
Medicine      & \citet{das2023twin}                    & Longitudinal Visit Events     & Baseline Clinical Characteristics                 & VAE                             & Risk Stratification \\
Medicine      & \citet{fang202342}                     & Chromosomal Abnormalities     & Metaphase Chromosome Images                       & Vision Transformers             & Population Proportions \\
Medicine      & \citet{lu2023multi}                    & Adverse Event Indicator       & Longitudinal Visit Data                           & Multi-Label Time Series GAN     & Population Proportions \\
Medicine      & \citet{mazumder2022synthetic}          & PPG Time Series, Binary CAD   & PPG Signals                                       & VAE                             & Regression Coefficients \\
Medicine      & \citet{pezoulas2020generation}         & Ten Physiologic Endpoints     & Demographics, Labs, Genetic Information           & Regression, Tree Ensembles      & Pearson Correlation Coefficient \\
Medicine      & \citet{rashid2019skin}                 & Synthetic Dermoscopic Images  & Real Dermoscopic Images, Lesion Annotations       & GAN + CNN Discriminator	     & Skin Lesion Classifications \\
Medicine      & \citet{romano2023exploring}            & Liver Cancer Risk             & SNPs                                              & Evolutionary Simulation Model   & Propensity Score Matching \\
Medicine      & \citet{shamsi2022karyotype}            & Chromosomal Abnormalities     & Karyogram Images                                  & Vision Transformers             & Population Proportions \\
Medicine      & \citet{shi2022generating}              & Treatment, Counterfactuals    & Demographics, Labs, Medications, Insurance Claims & GAN                             & Average Treatment Effect \\
Medicine      & \citet{yang2022identification}         & Metastatic Status             & Clinician Notes                                   & NLP                             & Regression Coefficients \\
Medicine      & \citet{zhang2023gan}                   & Binary Heart Disease Risk     & Demographics, Clinical Features                   & GAN                             & Regression Coefficients \\
Politics      & \citet{anastasopoulos2016photographic} & Racial Classification         & Images, Personal/District Demographics, Party     & CNNs                            & Regression Coefficients \\
Politics      & \citet{grundler2016democracy}          & Continuous Democracy Index    & Political Participation, Civil Liberty Metrics    & SVM                             & Mixed Effects Model Coefficients \\
Politics      & \citet{imai2016improving}              & Race (covariate)              & Surnames                                          & Bayesian Regression             & Regression Coefficients \\
Politics      & \citet{jamal2015anti}                  & Tweet Sentiment, Topic Focus  & Arabic-Language Tweets Referencing the U.S.       & LLMs                            & Univariate Associations \\
Politics      & \citet{king2013censorship}             & Censorship of Social Media    & Text in Social Media Post                         & LLMs                            & Bayesian Modeling, Hypothesis Testing \\
Politics      & \citet{martin2019local}                & Text from Media Outlets       & Text from Partisan Speeches                       & LDA Topic Model                 & Univariate Associations \\
Politics      & \citet{theocharis2016bad}              & Tweet Engagement Style        & Political Candidate and Party Characteristics     & Supervised Learning             & Multilevel Regression Coefficients \\
Public Health & \citet{lu2021beyond}                   & PM 2.5 Exposure               & Movement Data, PM 2.5 Concentration               & Random Forests                  & Univariate Associations \\
Sociology     & \citet{gebru2017using}                 & Motor Vehicle Category        & Google Street View Images                         & CNNs                            & Likelihood of Event \\
Sociology     & \citet{naik2017computer}               & Measure of Safety Perception  & Images of City Streetscapes                       & CNNs                            & Regression Coefficients \\
\bottomrule
\end{longtable}
\vspace{-2ex}
\noindent CAD: Coronary Artery Disease; CNNs: Convolutional Neural Networks; COPD: Chronic Obstructive Pulmonary Disease; DAC: Data Augmentation Classifier; DNNs: Deep Neural Networks; GAN: Generative Adversarial Network; GDP: Gross Domestic Product; LDA: Latent Dirichlet Allocation; LLMs: Large Language Models; MDD: Major Depressive Disorder; NLP: Natural Language Processing; PCA: Principal Components Analysis; PM 2.5: Fine Particulate Matter; PPG: Photoplethysmogram; SES: Socioeconomic Status; SNPs: Single Nucleotide Polymorphisms; SVMs: Support Vector Machines; VAE: Variational Autoencoder.

\end{landscape}

Our scientific target is a Euclidean functional, $\boldsymbol{\theta} = \Psi(P_{Y, \boldsymbol{X}})$, such as a mean, generalized linear model (GLM) coefficient, or causal contrast defined with respect to the joint law of $(Y, \boldsymbol{X})$. To motivate our problem, consider three estimators. First, the {\it oracle} estimator, $\hat{\boldsymbol{\theta}}^o$, uses all $n$ observations of $(Y, \boldsymbol{X})$. This is a hypothetical, as having complete $(Y, \boldsymbol{X})$ precludes our setup, but it serves as a benchmark. Second, the {\it classical} or complete-case estimator, $\hat{\boldsymbol{\theta}}^c$, uses only $\mathcal{L}$. Under i.i.d.~sampling, $\hat{\boldsymbol{\theta}}^c$ is statistically valid, but potentially inefficient, as it excludes $\mathcal{U}$. The third, {\it na{\"i}ve} estimator, $\hat{\boldsymbol{\eta}}$, uses all $n$ observations, but treats the predicted $\hat{Y}$ as observed $Y$. We caution against this, as it may lead to biased, anticonservative inference. We are also careful to denote the na{\"i}ve target by $\boldsymbol{\eta} = \Psi(P_{\hat{Y}, \boldsymbol{X}})$, rather than $\boldsymbol{\theta}$, as these parameters differ, except in the extreme case where $\hat{f}(\boldsymbol{Z}) = \mathbb{E}[Y \mid \boldsymbol{X}]$. This is not a new insight, as it is the motivation behind IPD and well-discussed in both~\citet{motwani2023revisiting} and~\citet{gronsbell2024another}, but we emphasize it here for clarity.

\subsection{The Problem}

AI/ML-generated data hold great promise for reducing costs and improving efficiency. Yet, treating predictions as observed data is not typically valid for statistical inference, as this can distort effect sizes and standard error estimates~\citep{ogburn2021warning}. Potential errors can enter an analysis through the same two mechanisms that underlie all of statistics: {\it bias} and {\it variance}. Namely, {\it bias} occurs when substituting $\hat{Y}$ for $Y$ systematically shifts the estimand so that na{\"i}ve procedures target $\boldsymbol{\eta} = \Psi(P_{\hat{Y}, \boldsymbol{X}})\neq\Psi(P_{Y, \boldsymbol{X}}) = \boldsymbol{\theta}$ or distorts the estimated relationships among variables. {\it Variance underestimation} occurs when uncertainty from the prediction model and the inherent variability of the true data are ignored or inadequately propagated. Crucially, these errors persist even when $\hat{Y}$ and $Y$ are highly correlated, and we emphasize that {\it high predictive accuracy does not ensure valid inference}. Before detailing these failure modes, we note that several common intuitions about prediction quality, error structure, and model confidence do not translate directly to valid inference. These themes reappear within each failure mode below.

\subsubsection{Bias}

Consider the bias of the na{\"i}ve estimator, which we obtain by simply substituting $\hat{Y}$ for $Y$. This shifts the target of inference from $\boldsymbol{\theta}$ to $\boldsymbol{\eta}$, so that the bias can be decomposed as
\[
    {\rm Bias}(\boldsymbol{\hat{\eta}}, \boldsymbol{\theta}) = \mathbb{E}[\boldsymbol{\hat{\eta}} - \boldsymbol{\theta}] = \mathbb{E}[\boldsymbol{\hat{\eta}} - \boldsymbol{\eta} + \boldsymbol{\eta} - \boldsymbol{\theta}]  = \underbrace{\mathbb{E}[\boldsymbol{\hat{\eta}} - \boldsymbol{\eta}]}_{\text{Estimation Bias}} + \underbrace{\mathbb{E}[\boldsymbol{\eta} - \boldsymbol{\theta}],}_{\text{Estimator Bias}}
\]
where $\boldsymbol{\theta}$ is our true scientific estimand, $\boldsymbol{\eta}$ is the target implicitly defined by substituting $\hat{Y}$ for $Y,$ and $\boldsymbol{\hat{\eta}}$ is our naive estimator. We can see from this decomposition that bias is introduced through two distinct mechanisms. {\it Estimation bias} concerns whether $\boldsymbol{\hat{\eta}}$ is consistent for the naive target, $\boldsymbol{\eta}$. {\it Estimator bias} concerns how far $\boldsymbol{\eta}$ is from the true target, $\boldsymbol{\theta}$. These two components can arise via different mechanisms and have different consequences for inference.

The first component, {\it estimation bias}, arises when $\boldsymbol{\hat{\eta}}$ does not estimate the naive target, $\boldsymbol{\eta}$, well. This can occur because $\hat{Y}$ is itself an estimate produced from finite data, with errors that depend on the model complexity, the chosen loss function, and stochasticity in the learning algorithm~\citep{hastie2009elements}. Several well-documented phenomena fall under {\it estimation bias}. First, a form of post-selection bias known as the `winner's curse' arises when the same data are used to both construct the prediction rule (e.g., in subgroup discovery) and to test for differences between the resulting groups~\citep{andrews2024inference, neufeld2022tree, zrnic2024flexible}. Here, `double-dipping' leads to overfitting that inflates the estimated effects relative to the true $\boldsymbol{\eta}$~\citep{athey2016recursive, fithian2014optimal}. Second, shrinkage induced by the choice of loss function creates directional biases, which can propagate downstream. For example, models trained under mean squared error (MSE) loss trade bias for reduced variance, which can cause these models to underestimate large values and vice versa. As a result, estimates computed using $\hat{Y}$ can be systematically attenuated relative to those based on $Y$~\citep{hastie2009elements, efron2020prediction, lee2025systematic}. We note that {\it estimation bias} tends to shrink as training data grow larger and more representative, and as the predictive models better approximate the true data generating mechanism. However, in small or moderate samples, or in settings of heterogeneity, there is `no free lunch'~\citep{mani2025no}.

The second component, {\it estimator bias}, arises because, even with infinite data, the naive estimator converges to the wrong estimand. The intuition is that by replacing the true outcome, $Y$, with a prediction, $\hat{Y}$, you are changing the underlying data-generating structure that defines the target parameter, $\Psi(P_{Y, \boldsymbol{X}})$. Here, {\it high predictive accuracy} tells you how {\it close} $\hat{Y}$ is to $Y$ on average, but valid inference depends on which part of $Y$ matters for the parameter, $\Psi$. Predictive models are typically trained to approximate $Y$ pointwise, rather than preserve its specific structure, such as its functional form, its true underlying variability, and its dependence on $\boldsymbol{X}$. But the prediction problem is fundamentally overdetermined, as there are many different functions of $\boldsymbol{X}$ that yield low prediction error, with only a subset of them preserving the specific structural relationship between $Y$ and $\boldsymbol{X}$ that defines the target parameter, $\Psi(P_{Y, \boldsymbol{X}})$. This {\it structural distortion} is an example of {\it estimator bias}. Specifically, $\hat{f}$ can introduce nonlinearities, interactions, and artifacts not present in $Y$. Using $\hat{Y}$ to fit downstream models changes the {\it shape} of the relationships between the outcome and covariates, as the downstream model `sees' the structure of $\hat{Y}$, rather than $Y$. Two practical drivers of {\it structural distortion} in this context are particularly relevant. First, {\it domain shift}, occurs when $\hat{f}$ is applied to a population that is out of distribution relative to the training data, changing error patterns across subpopulations and creating spurious or attenuated associations in the analytic sample~\citep{subbaswamy2021evaluating, pooch2020can, stacke2020measuring, hoffman2024some}. Second, {\it model overfitting and opacity} exacerbate the problem. Over-flexible predictors can produce seemingly accurate, but unstable signals that do not generalize, while measures of uncertainty are misaligned with the inferential target and can subsequently mask distortions~\citep{hawkins2004problem, belkin2019reconciling, bhatt2021uncertainty, kompa2021second, cao2023extrapolation}. 

\citet{ogburn2021warning} further show that replacing observed outcomes with predicted or surrogate values can have unexpected consequences, especially when complex causal relationships are involved. Specifically, using predicted outcomes as conditioning variables can create a collider bias. This occurs when two otherwise independent factors both influence the predicted outcome, and conditioning on that shared outcome induces a spurious correlation between them. \citet{ogburn2021warning} demonstrate scenarios where this bias arises under both the null and alternative hypotheses. Another related issue arises when we have predictions generated from a model that does not include our features of interest, $\boldsymbol{X}$, but instead uses correlated surrogate features, $\boldsymbol{Z}$. Such predictions may be highly accurate in terms of mean squared error (MSE) due to the strong correlation between $\boldsymbol{Z}$ and $Y$, however, any estimated associations can be distorted in both magnitude and direction. Thus, even very accurate predictions can distort the relevant signal for estimating $\Psi$ As a result, the naive estimator converges to $\Psi(P_{\hat{Y}, \boldsymbol{X}})$, which generally differs from $\Psi(P_{Y, \boldsymbol{X}})$, even when the correlation between $\hat{Y}$ and $Y$ is high~\citep{hastie2009elements, efron2020prediction, lee2025systematic}. 

Lastly, it is important to note that no amount of additional predicted data can add information when the parameter is already {\it well-specified}, meaning that it does not depend on the marginal distribution of $\boldsymbol{X}$. \citet{xu2025unified} formalize this distinction under a unified framework for semi-supervised learning, which generalizes results for linear models~\citep[e.g., see][]{kawakita2013semi, buja2019models, song2024general} to any arbitrary inferential problem. 

\subsubsection{Variance Underestimation}

{\it Variance underestimation} completes the picture.  {\it Variance underestimation} follows because naive analyses treat $\hat{Y}$ as observed and fixed, which ignores variability from both the prediction model and the true underlying data generating mechanism. Here, two components of variance are typically lost: (i) residual noise around the fitted function, $\hat{f}(\boldsymbol{Z})$, and (ii) finite-sample uncertainty from estimating $\hat{f}$ itself. From a missing data perspective, replacing $Y$ with a realization of $\hat{Y}$ is like single imputation, which underestimates uncertainty about the underlying data generating mechanism, a form of measurement error~\citep{little2019statistical, rubin1991multiple}. A common example is when latent class membership is predicted upstream (e.g., by taking the modal class probability), and reified into a downstream model despite uncertainty in the class membership~\citep[e.g., see][]{abdar2021review, elliott2020methods}. 

In modern predictive models, both (i) and (ii) are often {\it heteroskedastic} and {\it dependent} in ways classical models do not capture. For one, prediction errors can vary systematically in the feature space and can display dependence induced by shared inputs, smoothing, or algorithmic regularization~\citep{csahin2025unlocking, gelfand2015understanding, hooker2021bridging, wager2018estimation, barber2021predictive}. Further, error distributions are frequently non-Gaussian, asymmetric, or heavy-tailed~\citep{hastie2009elements}. Ignoring this leads to underestimated standard errors and anticonservative inference, which is exacerbated by the empirical overconfidence of many black-box predictors~\citep{guo2017calibration, ovadia2019can, aliferis2024overfitting, groot2024overconfidence}. 

At the same time, there are cases where predicted data can improve precision. Namely, when the prediction residuals carry information about $Y$ beyond what $\boldsymbol{X}$ encodes, and when the error variance is small and stable, prediction-based inference procedures can be more efficient than the classical, i.e., complete-case analysis~\citep{gronsbell2024another}. These efficiency gains, however, are conditional and fail when the residual correlation is weak, the error variance is large or heterogeneous, or when the estimand does not benefit from the unlabeled marginal distribution~\citep{xu2025unified}.

\section{Example Cases Where Inference Can Go Wrong}

\subsection{An Illustrative Example}

We take a stylized linear example to make the these sources of error concrete. Let $\boldsymbol{Z} = (Z_1, \ldots, Z_{10})$ with $Z_j \stackrel{\rm i.i.d.}{\sim}\mathcal{N}(0,1)$ and

\[
    Y=\sum_{j=1}^{10} Z_j + \varepsilon, \quad \varepsilon \sim \mathcal{N}(0,1).
\]

The inferential goal is the marginal slope of $Z_1$ on $Y$. We train random forests on an independent sample of size $n_t = 1,000$: $\hat{f}^{(1)}$, which uses all ten predictors, $\hat{f}^{(2)}$, which uses $(Z_1, Z_2)$, $\hat{f}^{(3)}$, which uses $(Z_2, \ldots, Z_{10})$ and {\it excludes} $Z_1$, and $\hat{f}^{(4)}$, which uses only $(Z_2, Z_3)$, again {\it excluding} $Z_1$. Here, these settings are meant to show cases when either/both bias and variance underestimation arise.

\begin{figure}[!ht]
    \centering
    \includegraphics[width=\linewidth]{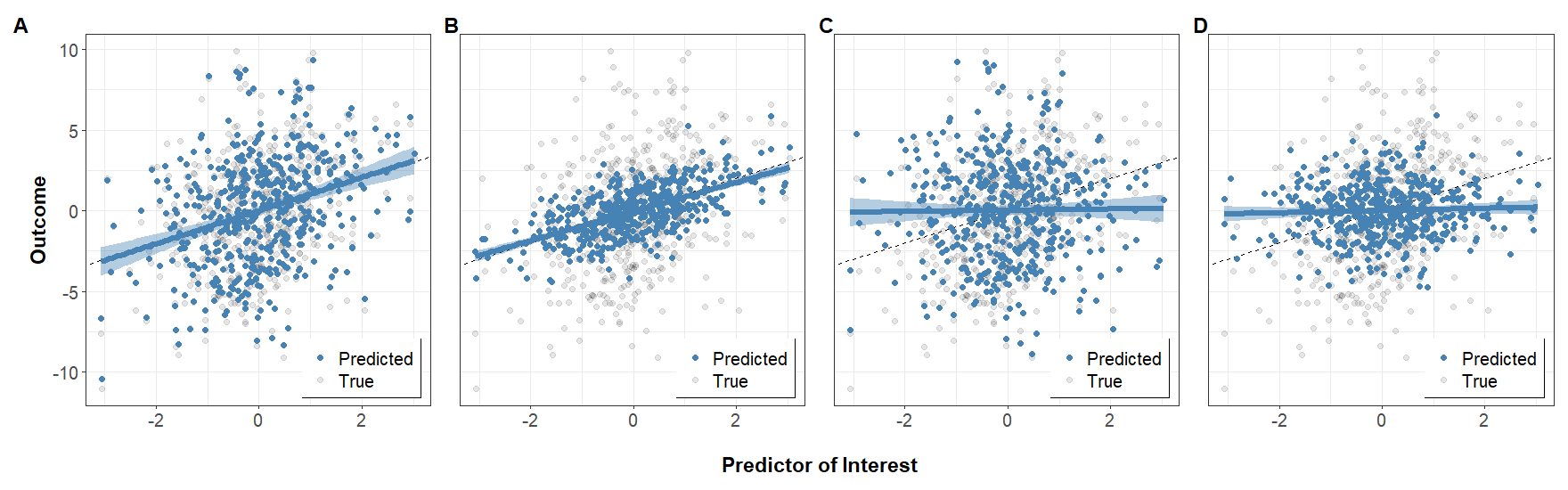}
    \caption{Illustrative example of bias and variance when regressing a predicted outcome on the predictor of interest. Each panel plots the outcome vs.~$Z_1$: gray points are the true $Y$, blue points are the predicted $\hat{Y}$ from rules trained on an independent sample. The black dashed line is the true slope, the blue line and ribbon are the fitted line and its standard error from regressing $\hat{Y}$ on $Z_1$. Prediction rules use different feature sets: (A) all ten features, (B) $Z_1, Z_2$, (C) $Z_2, \ldots, Z_{10}$ (excluding $Z_1$), (D) $Z_2, Z_3$ (excluding $Z_1$).}
    \label{fig:illustration}
\end{figure}

We then apply each rule to an analytic sample of size $n = 1,000$ to form $\hat{Y}^{(m)} = \hat{f}^{(m)}(\boldsymbol{Z})$, and we regress each $\hat{Y}^{(m)}$ on $Z_1$ (Figure \ref{fig:illustration}). When all predictors are used to construct the predicted outcome ($\hat{Y}^{(1)}$, Panel A), the bias is near zero, but the variability around the regression line is slightly reduced because the outcomes have been replaced by fitted conditional means. When the rule includes the predictor of interest but omits most others ($\hat{Y}^{(2)}$, Panel B), the slope remains near the truth, while the residual variance is badly underestimated, producing anticonservative intervals. When the rule excludes the predictor of interest entirely, the estimated slope is near zero despite high overall predictive fit with other covariates ($\hat{Y}^{(3)}$, Panel C). And when the rule includes neither the predictor of interest nor the remaining drivers ($\hat{Y}^{(4)}$, Panel D), both bias and under-variance appear. Predictive accuracy can be high in all four cases, yet the inferential behavior differs in ways explained exactly by issues of bias propagation and variance underestimation. For a more detailed derivation of the analytic biases and standard errors for this example, please see Appendix \ref{sec:A}.

\subsection{A Simulation-Based Case Study: Estimating Voter Turnout by Race}

Our illustrative example shows how na{\"i}ve predictions can bias inference, even when the predictions correlate strongly with the truth. We now examine the same phenomena in a real application: estimating voter turnout. Specifically, to estimate racial disparities in voter turnout, researchers often predict an individual's race in voter files using methods like Bayesian Improved Surname Geocoding~\citep[BISG, see][]{ImaiKhannaReplication}. When these predicted groups are then used for downstream inference, they can induce an {\it ecological fallacy-like} error. Classical ecological fallacy arises when aggregated associations are misinterpreted as individual relationships. Here the mechanism is inverted but analogous: individual race is {\it predicted} from aggregate features (here surname and geography), then treated as ground truth and aggregated to estimate group-specific voter turnout. This two-step pipeline can induce estimand mismatch and alter the association structure (bias), as well as omit uncertainty from the prediction step (variance underestimation).

We use Florida voter registration data~\citep{ImaiKhannaReplication}, which contain both self-reported race (gold-standard labels) and surname-based race probabilities from the U.S.~Census Surname List~\citep{USCensus2007Surname}. Our goal is to estimate the probability of voting in the 2008 general election conditional on race, i.e., $\Pr({\rm Voted} = 1 \mid {\rm Race} = r)$, for each of five groups: White, Black, Hispanic, Asian, and Other Race. To explore how well different approaches recover these probabilities, we compare two inferential strategies: (1) using the gold-standard (true) labels and (2) na{\"i}vely using predicted surrogates. For the gold-standard analysis, we restrict to individuals with observed race and directly compute empirical turnout rates within each group. For comparability with other methods, we also estimate $\Pr({\rm Race} = r \mid {\rm Voted} = 1)$ via a logistic regression to recover
\[
    \Pr({\rm Voted} = 1 \mid {\rm Race} = r) = \frac{\Pr({\rm Race} = r \mid {\rm Voted} = 1) \times \Pr({\rm Voted} = 1)}{\Pr({\rm Race} = r)}.
\]
This serves as our benchmark. For the na{\"i}ve approach, we assign each voter to their most probable race based on the surname-derived probabilities from the U.S. Census Surname List. We then fit a logistic regression to estimate $\Pr({\rm Race} = r \mid {\rm Voted} = 1)$ using these predicted groups, and we use Bayes' Rule to estimate $\Pr({\rm Voted} = 1 \mid \widehat{{\rm Race}} = r)$. While surname is suggested to be a strong predictor of race, this approach is expected to be biased when used for downstream inference.

Table \ref{tab:turnout-stratified} reveals several key insights about the interaction between prediction quality and inferential validity, namely that high prediction accuracy does not ensure valid inference. As shown, the na{\"i}ve method exhibits non-trivial bias even when classification performance is strong. For instance, the ``White'' group has an 80\% classification accuracy and AUC of 0.83, yet underestimates turnout by nearly one percentage point. Similarly, the ``Other'' group achieves 96\% accuracy, but the na{\"i}ve estimate deviates by -1.71\%. This highlights that high predictive accuracy on individual quantities is insufficient for valid inference on population-level conditional quantities.

\begin{table}[!ht] 
\centering 
\caption{Voter turnout probability by race (true vs.~na{\"i}ve). Error uses absolute deviation from truth; higher Accuracy and AUC are better (blue = better, yellow = worse).}
\label{tab:turnout-stratified} 
\vspace{0.5ex} 
\begin{tabularx}{\textwidth}{l R R R R R} 
\toprule 
\textbf{Race} & \textbf{True} & \textbf{Naive} & \textbf{Naive Err} & \textbf{Accuracy} & \textbf{AUC} \\ 
\midrule 
White & 0.7488 & 0.7392 & \shadeabserr{-0.0096}{0.0175} & \shadefromto{0.8013}{0.8013}{0.9853} & \shadefromto{0.8300}{0.5242}{0.9341} \\ 
Black & 0.7155 & 0.7145 & \shadeabserr{-0.0010}{0.0175} & \shadefromto{0.8812}{0.8016}{0.9853} & \shadefromto{0.8479}{0.5242}{0.9341} \\ 
Hispanic & 0.6282 & 0.6336 & \shadeabserr{ 0.0054}{0.0175} & \shadefromto{0.9422}{0.8016}{0.9853} & \shadefromto{0.9336}{0.5242}{0.9341} \\ 
Asian & 0.6100 & 0.6069 & \shadeabserr{-0.0031}{0.0175} & \shadefromto{0.9851}{0.8016}{0.9853} & \shadefromto{0.8547}{0.5242}{0.9341} \\ 
Other & 0.6384 & 0.6213 & \shadeabserr{-0.0171}{0.0175} & \shadefromto{0.9609}{0.8016}{0.9853} & \shadefromto{0.5244}{0.5242}{0.9341} \\ 
\bottomrule 
\end{tabularx} 
\end{table}

\subsection{A Public Health Case Study: Measuring Obesity.}

Further, predictions do not have to be complicated to cause issues. We now present a simple example where surrogate measures of adiposity can lead to substantively different conclusions in public health research. Body mass index (BMI, kg/m$^2$) is a deterministic prediction of unobserved percent body fat. It is inexpensive and standardized, but it neither separates fat from lean mass nor does it capture an individual's fat distribution. As a result, it can under- or over-estimate true adiposity in key subgroups~\citep[e.g., biological males vs.~females, athletes, older adults with sarcopenia, see][]{jeong2023different, razak2007defining, bogin2012body}. BMI is often defended for use in population-level inference on the premise that individual misclassifications `average out'~\citep{gutin2018bmi}, yet this reasoning masks the problem we are trying to illustrate: BMI a prediction of adiposity, and its systematic errors need not vanish with aggregation~\citep{rothman2008modern, visokay2025measure}.

Following \citet{visokay2025measure}, we compare BMI to waist circumference (WC) and dual-energy X-ray absorptiometry (DXA) percent body fat. WC is a simple proxy for central adiposity and cardiometabolic risk~\citep{ness2008waist, huxley2010body}, but, like BMI, it is not a direct measure and is sensitive to posture and body habitus. DXA provides a gold-standard assessment of total and regional fat and lean mass, with validated percent fat thresholds~\citep{bazzocchi2016dxa, shepherd2017body, potter2025defining}, but it is costly to measure.

\begin{figure}[!ht]
    \centering
    \includegraphics[width=\linewidth]{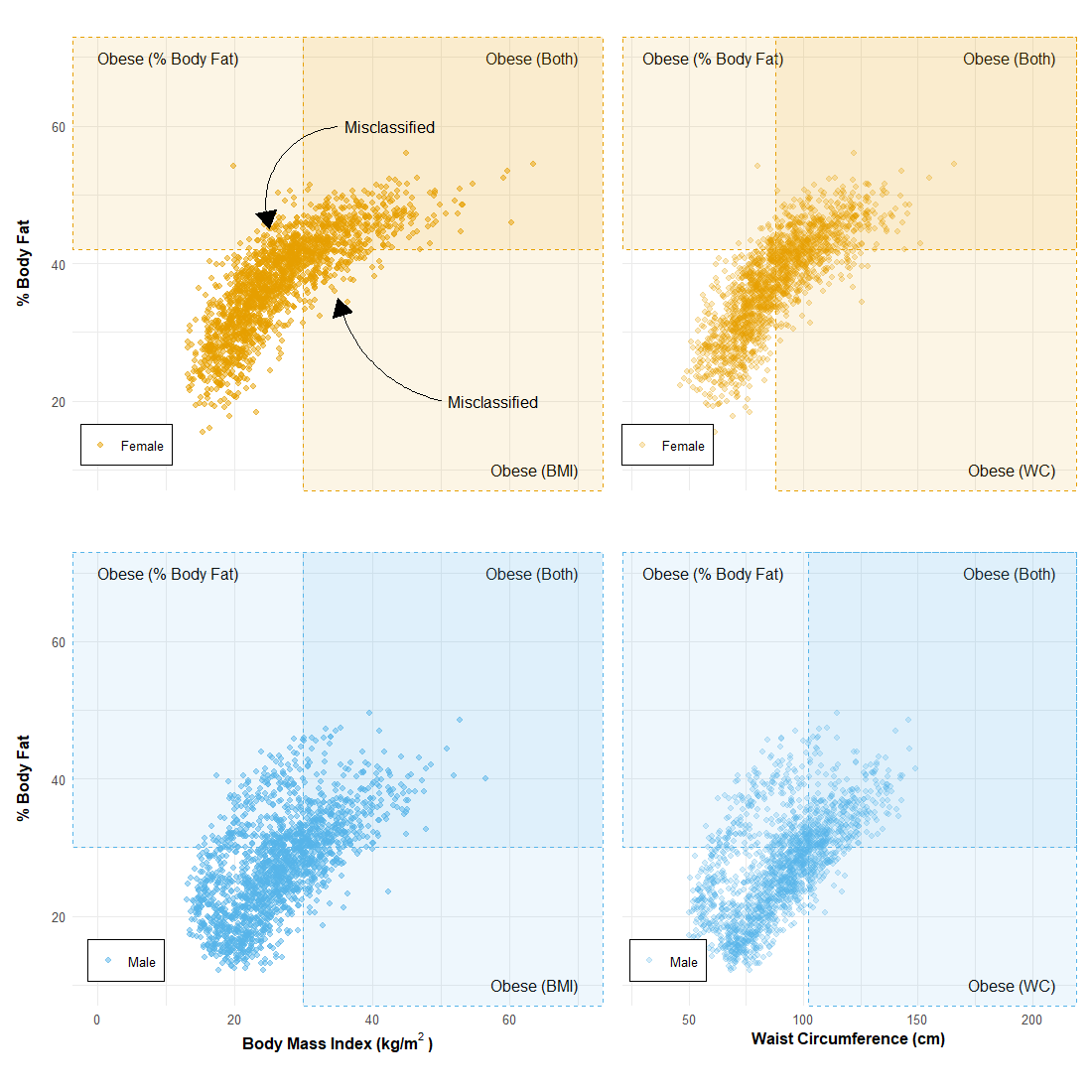}
    \caption{DXA percent body fat vs.~two proxy measures. Shaded regions within dashed lines mark obesity thresholds for each measure, showing misclassification under either proxy measure (top-left and bottom-right regions of each panel) and the overlap between them (top-right) for BMI (kg/m$^2$, left) and waist circumference (cm, right). This misclassification differs for females (top, yellow) versus males (bottom, blue).}
    \label{fig:bmi_scatter}
\end{figure}

We use the National Health and Nutrition Examination Survey (NHANES), a nationally representative survey with standardized exams~\citep{batsis2016diagnostic, johnson2013national}. DXA was collected through 2017-2018 but suspended during the COVID-19 pandemic, so later waves (e.g., Aug 2021-Aug 2023) lack DXA measurements~\citep{paulose2021national}. Figure~\ref{fig:bmi_scatter} compares DXA percent fat to BMI and WC. Both proxies correlate strongly with DXA, but the distribution of percent body fat is visibly bimodal, indicating the different body composition patterns between males and females. Further, threshold-based obesity classifications (BMI: > 30 kg/m$^2$, WC: $\geq$ 102 cm in males, $\geq$ 88 cm in females, DXA: > 30\% in males, > 42\% in females) disagree for a nontrivial share of participants and differ by sex, consistent with subgroup calibration error.

\begin{figure}[!ht]
    \centering
    \includegraphics[width=\linewidth]{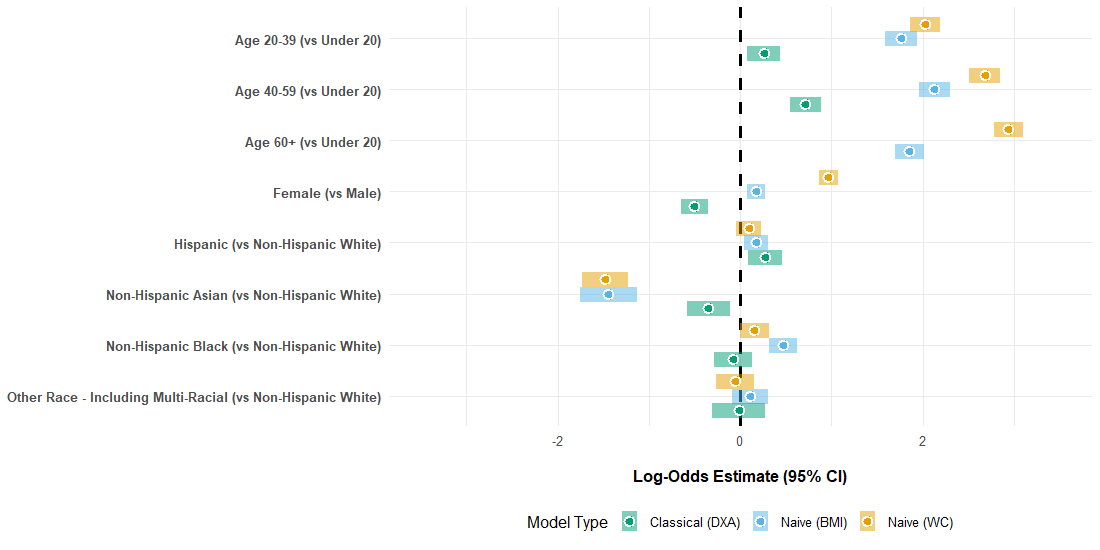}
    \caption{Coefficient (log-odds) estimates and 95\% confidence intervals for logistic regressions of obesity on certain demographic risk factors (age, sex, and race). Obesity is a binary outcome defined based on pre-specified thresholds for three continuous measures of adiposity: dual-energy X-ray absorptiometry (DXA)-based adiposity \% body fat (green), body mass index (BMI; kg/m$^2$, blue), and waist circumference (WC; cm, yellow). }
    \label{fig:bmi_forest1}
\end{figure}

To examine how this impacts downstream inference, we model obesity as a binary outcome using each measure (DXA, BMI, WC) in separate logistic regressions on age, sex, and race. The DXA-based obesity model serves as our oracle, while the BMI and WC models are two na{\"i}ve approaches. Figure~\ref{fig:bmi_forest1} shows that effect estimates differ in both magnitude and/or sign across subgroups. This is structural distortion: the association structure induced by BMI/WC is not the association with true adiposity. At the same time, average misclassification relative to DXA produces bias propagation in group contrasts, and treating BMI/WC as if observed measurements underestimates variance. These takeaways echo our the earlier examples: high correlation does not guarantee valid inference unless bias and uncertainty from the prediction step are explicitly propagated.

Lastly, note that in all three examples, each failure mode could be diagnosed using gold-standard labeled data by comparing $Y$ and $\hat{Y}$ directly via calibration summaries and by performing side-by-side regressions of $Y$ and $\hat{Y}$ on $\boldsymbol{X}$. It is important for researchers to be clear about their inferential target, $\Psi(P_{X,Y})$, as this may not be what they are actually measuring, the labeled fraction of their data $r = n_\ell/n$, and which quantities are being predicted. These checks, and the necessity of a labeled subset, $\mathcal{L}$, naturally motivate the recent assumption-lean methods for inference with predicted data that we will now review in the next section.

Ultimately, every inferential error in the naive analysis reflects the same two axes of statistical error: {\it bias}, when the use of predicted values change what is being estimated, and {\it variance}, when uncertainty from both the model and the data are not accounted for. The challenge, and opportunity, of conducting inference with predicted data is to quantify and correct for both sources of error simultaneously.

\section{Emerging Solutions and the Road Ahead}
\label{sec:3}

The challenge of conducting valid {\it inference with predicted data} (IPD) is that treating predictions as observed data can produce biased estimates, underestimated variance, and spurious, attenuated, or even reversed associations, often unpredictably. A natural solution leverages a small labeled set with gold-standard $(Y, \boldsymbol{X})$ to calibrate inference in a larger sample containing AI/ML-generated data. This is increasingly relevant with the rise of `highly accurate' predictive models in fields such as medicine, economics, and AI itself. IPD methods correct for bias and additional uncertainty from the prediction step, allowing for increased sample sizes while still ensuring valid inference. These techniques further emphasize {\it assumption-lean} frameworks, which treat the predictive models as `black-boxes.' Despite minimal assumptions, many of these approaches guarantee unbiasedness and correct coverage probabilities, ensuring reliable inference regardless of prediction quality.

In the past five years, the availability of methods for IPD has expanded rapidly. We use `IPD' as an umbrella term for inference tasks, including estimating associations, testing hypotheses, and constructing confidence intervals, when key variables are supplied by predictive models, rather than directly observed. When referring to particular approaches (e.g., post-prediction inference, prediction-powered inference), we use their names. Our goal is not to provide an exhaustive or to adjudicate among competing methods. Instead, given the pace of development, we aim to highlight several influential contributions from the recent literature. Further, despite the novelty of IPD, it has deep roots in fundamental statistical theory, which we will later underscore.

\subsection{A Brief Overview of Recent Contributions}

\citet{wang2020methods} provided a formal treatment of this problem with {\it post-prediction inference} (PostPI), a correction procedure for inference on predicted outcomes. PostPI fits into a natural pipeline: (i) train a prediction model upstream, (i) collect gold-standard pilot data to learn the relationship between predicted and true outcomes, and (iii) apply this relationship to future studies with missing outcomes. A key insight was that the relationship between $\hat{Y}$ and $Y$ can often be modeled via a low-dimensional calibration curve. This `relationship model' can be applied downstream to correct bias and inflate standard errors, improving over the na{\"i}ve approach. However, \citet{motwani2023revisiting} identified that the validity of PostPI relies on the assumption that the predictions accurately reflect the conditional relationship between the true outcome and covariates. \citet{salerno2025moment} have since proposed a simple, moment-based extension to the PostPI procedure, which relaxes the original method's assumptions, but this highlights the need for more robust inferential methods in this space.

In 2023, \citeauthor{angelopoulos2023prediction} introduced {\it prediction-powered inference} (PPI) as a more general and theoretically rigorous framework for IPD. Like PostPI, PPI leverages a small, labeled dataset to calibrate inference in a larger, unlabeled dataset. PPI forms the na{\"i}ve estimate in the unlabeled set, which it {\it rectifies} by modeling the residual difference between the predicted and observed outcomes in the labeled set. The key theoretical innovation is PPI's validity guarantee, which holds while making {\it no assumptions} about the upstream AI/ML model. Namely, the estimate is unbiased, Type I error is always controlled, and confidence intervals achieve nominal coverage, even when predictions are poor. Simultaneously, `better' predictions naturally yield narrower intervals and increased statistical power. This assumption-lean approach differs from traditional semi-supervised methods, which often rely heavily on stronger distributional or model-based assumptions.

While PPI is a significant advancement due to its inferential \textit{validity}, an important line of research has focused on improving its statistical {\it efficiency} and computational {\it simplicity}~\citep{angelopoulos2024note}. To this end, \citet{angelopoulos2023ppi++} introduced PPI++, which provides more efficient standard error estimation (the first `+'), and optimal weighting to minimize variance ({\it power-tuning},  the second `+'). A notable strength is PPI++'s data-driven adaptivity: when predictions are informative, it leverages them to increase precision; when they are not, it reduces to the classical estimator. However, as discussed in \citet{angelopoulos2024note}, while PPI++ significantly enhances practical efficiency, it does not attain semiparametric efficiency. This is an active area of research, as discussed in \citet{gronsbell2024another}, \citet{xu2025unified}, and \citet{ji2025predictions}.

More recent developments, such as PoSt-Prediction Adaptive inference~\citep[PSPA;][]{miao2023assumption}, Prediction De-Correlated inference~\citep[PDC;]{gan2024prediction}, and the `CC' method proposed by \citet{gronsbell2024another}, have further generalized and expanded this framework. Like PPI(++), PSPA is {\it assumption-lean}, on the prediction model, and {\it data-adaptive}, optimally weighting the predictions. It uses a vector of weights estimated from the labeled data, ensuring both bias correction and variance reduction for each parameter estimated, and is broadly applicable to common inferential targets that can be written as solutions to estimating equations~\citep{miao2023assumption}. Similarly, PDC is a safe and assumption-lean method that proposes a matrix-valued augmentation, yielding a one-step estimator with improved efficiency~\citep{gan2024prediction}. \citet{gronsbell2024another} revisit \citet{chen2000unified} in the context of IPD and derive an even more efficient estimator by adaptively weighting and augmenting with the full dataset. They also draw important connections to augmented inverse probability weighting and show how IPD methods can relax the common assumption that $Y$ is {\it missing completely at random} (MCAR) to {\it missing at random} (MAR), aligning with classical results from \citet{robins1994estimation} and subsequent work in semiparametric inference.

\begin{figure}[!ht]
    \centering
    \includegraphics[width=\linewidth]{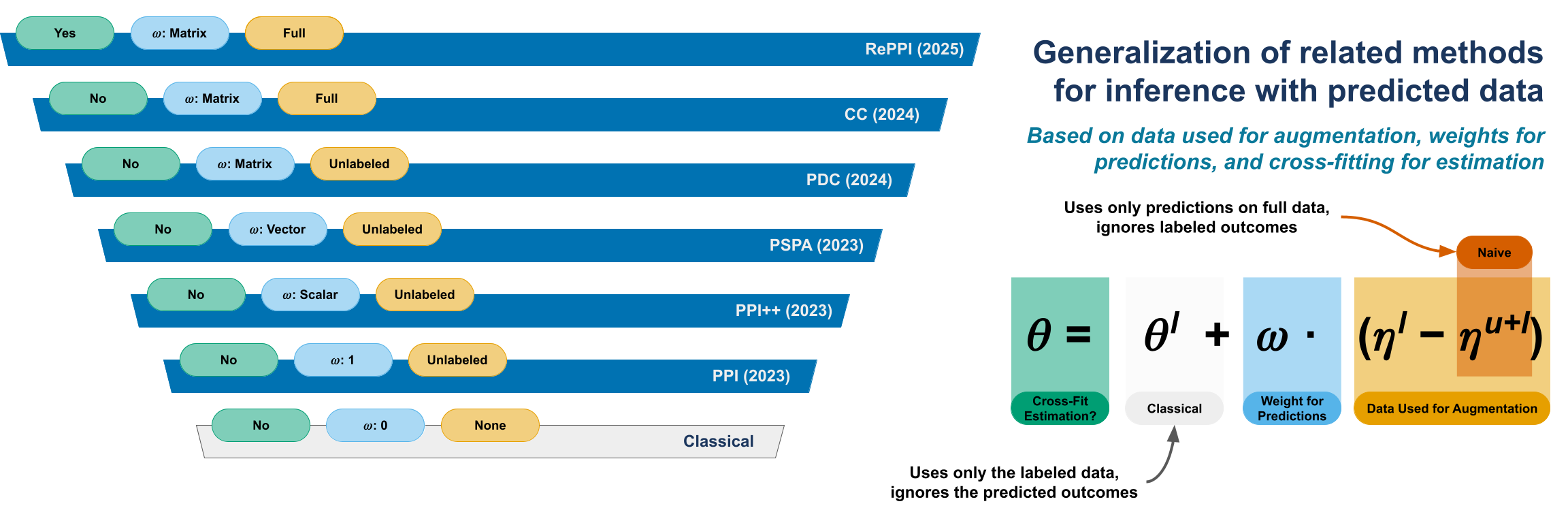}
    \caption{Comparison of related methods for inference with predicted data. Methods included are RePPI~\citep{ji2025predictions}, CC~\citep{gronsbell2024another},  PDC~\citep{gan2024prediction}, PSPA~\citep{miao2023assumption}, PPI++~\citep{angelopoulos2023ppi++}, and PPI~\citep{angelopoulos2023prediction}.}
    \label{fig:ipd_methods}
\end{figure}

\citet{ji2025predictions} develop a unifying framework that connects and generalizes many existing methods, including PPI and PPI++, PSPA, PDC, and CC, and establish connections to surrogate outcomes and semiparametric theory with Recalibrated PPI (RePPI). Their estimator is given by
\begin{equation}
    \label{eq:unifying}
    \hat{\theta} = \mathop{\arg \min}\limits_{\theta} \frac{1}{n} \sum_{i = 1}^n \ell_{\theta} \left(\boldsymbol{X}_i, Y_i\right) - \left\{\frac{1}{n} \sum_{i = 1}^n g_{\theta} \left[\boldsymbol{X}_i, f(\boldsymbol{Z}_i)\right] - \frac{1}{N} \sum_{i = n + 1}^{n + N} g_{\theta} \left[\boldsymbol{X}_i, f(\boldsymbol{Z}_i)\right]\right\},
\end{equation}
where $\ell_{\theta}(\cdot)$ is the target {\it loss} and $g_{\theta}(\cdot)$ is a method-specific function they call the {\it imputed loss}. Under this framework, we can see explicit connections to the previously-discussed methods. Taking $g_{\theta} = 0$ reduces to the `classical' estimator. PPI sets $g_{\theta} = \ell_{\theta}$ and can be thought of as a special case of \citet{tang2012efficient} with a single imputation from $\hat{f}$ and constant $(\tfrac{n}{n + N})$ propensity score. PPI++ and PSPA choose $g_{\theta}$ so that $\nabla g_{\theta} = \hat{M}\nabla \ell_{\theta}$, where $\hat{M}$ is estimated to minimize the asymptotic variance of $\hat{\theta}$ within a class $\mathcal{M}$ (e.g., $\mathcal{M}$ is the set of scaled identity matrices or diagonal matrices for PPI++ and PSPA, respectively). This formulation is also thematically similar to that of \citet{schmutz2022don}. PDC generalizes the work of \citet{song2024general} and takes $\nabla g_\theta = \gamma\hat{M}(\theta)h_\theta$, with scalar $\gamma$, $p \times q$ matrix $\hat{M}$, and a given $h_\theta : \mathbb{R}^p \to \mathbb{R}^q$ (here, $h_\theta = \nabla \ell_\theta$). CC take $\ell_{\theta}$ to be the squared loss for linear regression, which is related to both \citet{chen2000unified} and \citet{neyman1959optimal} (see Figure \ref{fig:ipd_methods}). In general, \citet{ji2025predictions} show that for convex $\ell_{\theta}$, the choice of $\nabla g_{\theta}$ equal to some $\psi_\theta(\boldsymbol{X}, f(\boldsymbol{Z}))$ is essentially the augmented estimator of \citet{robins1994estimation}. This establishes a formal connection between IPD and surrogate outcomes. RePPI learns the optimal {\it imputed loss} to further calibrate predicted outcomes via cross-fitting~\citep{chernozhukov2018double}. They show that RePPI is always more efficient than `single imputation-based' IPD methods and can reach the semiparametric efficiency bound when $g_\theta$ is estimated consistently. This moves IPD toward a unified, robust, and efficiency-aware framework that is grounded in semiparametric theory.

\subsection{Practical and Focused Areas.} IPD methods have advanced across several application areas, with methods tailored to each field's unique needs. One example is active experimentation. In resource-constrained settings, active labeling can dramatically reduce cost while preserving inferential precision. \citet{zhu2024doubly} propose doubly-robust self-training, which iteratively refines both a prediction model and an inference estimator under a semi-supervised regime. \citet{ao2024prediction} introduce prediction-guided active experiments, which select experimental units for labeling to reduce inferential uncertainty. Semi-supervised risk control adapts PPI's framework to tune the error rate of prediction rules.~\citep{einbinder2024semi}. \citet{angelopoulos2025cost} study cost-optimal annotation by allocating a given budget between weak and strong raters to maximize efficiency. Further, as large language models proliferate, efficient evaluation becomes critical. \citet{chatzi2024prediction} present prediction-powered ranking, which uses human-labeled data to calibrate model-generated quality scores. \citet{fisch2024stratified} extend PPI to stratified settings with hybrid language model evaluation, combining human judgments and model-predicted metrics across predefined strata. In fairness auditing, the process of evaluating a model's performance across subpopulations, ideas from IPD and semi-supervised learning can help when inequity is a concern~\citep{gao2025reliable}.

In causal, surrogate, and clinical trial settings, IPD is being used to extend transportability and surrogate endpoint ideas. \citet{demirel2024prediction}'s prediction-powered generalization of causal inferences supplements randomized control trial (RCT) data with predictions from observational studies. This facilitates generalization when the observational study is `high-quality' and remains robust when it is not. \citet{poulet2025prediction} use digital twins to run RCTs with fewer controls, reducing the sample size required to maintain the same power as a standard RCT. \citet{gradu2025valid} demonstrate valid inference after causal discovery, which corrects naive combinations of causal discovery algorithms and inference methods that would otherwise have high false positive rates due to `double-dipping.' RePPI unifies these threads, learning an optimal `imputed loss' and employing cross-fitting to safeguard against overfitting in small labeled samples~\citep{ji2025predictions}.

In genomics, \citet{miao2024valid} and \citet{mccaw2024synthetic} apply IPD to machine learning–assisted GWAS, where AI/ML is used to generate phenotypes for genetic discovery. Their methods, Post-Prediction GWAS (POP-GWAS) and Sythetic Surrogates (SynSurr), correct for errors arising from the use of AI/ML-imputed phenotypes. \citet{miao2024valid}, \citet{mccaw2024synthetic}, and \citet{mukherjee2024causal}'s complementary work show that the true positive rate (TPR) and false discovery rate (FDR) depend on the causal relationships among the inputs to the prediction model, the true phenotypes, and the environment. Namely, they show that  false discoveries occur when the predicted phenotype is based on downstream biomarkers and that an AI/ML-assisted GWAS's power depends on the heritability of the true phenotype and its correlation with the predictions.

In experimental and designed settings, design-based semi-supervised learning (DSL) prescribes a sampling design under a missing at random (MAR) assumption and then applies IPD corrections to the resulting pseudo-observations, ensuring that both bias and variance are properly accounted in a doubly-robust manner~\citep{egami2023using}. While \citet{egami2023using} focus on structured surrogates that have been predicted from text-based auxiliary data (e.g., by a large language model), \citet{rister2025correcting} adapt DSL to correct measurement errors in AI-assisted labeling for image analysis. \citet{kluger2025prediction} further extend these `predict-then-debias' estimators to give valid bootstrap confidence that apply when the complete data arise from weighted, stratified, or clustered (i.e., nonuniform) sampling designs. \citet{waldetoft2025prediction} also develop IPD estimators for finite-population statistics, with a focus on highly imbalanced textual data for hate crime estimation. In prospective designs, researchers can go one step further with  AI/ML-assisted data collection. Assuming a fixed budget, \citet{zrnic2024active} propose a methodology that uses AI/ML to identify which data points would be most beneficial to label.

IPD has also been extended to Bayesian, federated, and simulation-based settings. \citet{hofer2024bayesian} propose Bayesian PPI, which incorporates Bayesian-style design of new PPI-based estimands and posterior uncertainty via credible intervals, while \citet{o2025ai} exploit the inherent randomness of generative AI responses to construct prior distributions. \citet{luo2024federated} have proposed federated PPI, which enables decentralized inference across institutions without sharing raw data. \citet{li2024prediction}'s prediction-enhanced Monte Carlo uses AI/ML predictions as control variates to reduce simulation variance. Further, \citet{testa2025semiparametric} introduce a semiparametric framework that remains valid under distribution shift, while \citet{li2025statistical} apply IPD to inference under performativity, where predictions themselves can alter the data-generating process.

IPD has also been extended to meet several highly specialized methodological challenges. For example, \citet{song2025sparse} combine sparse semi-supervised regression with IPD adjustments to perform reliable feature selection in high-dimensional settings, while \citet{li2025prediction} develop prediction-powered adaptive shrinkage estimation, using model-generated predictions to guide principled shrinkage of noisy estimates. In randomized text-based trials, \citet{mozer2025more} demonstrate that IPD corrections can improve inference when outcomes are imputed via natural language models. \citet{gu2024local} introduce local PPI, which applies polynomial regression in localized regions of the covariate space to capture non‐stationary prediction error. \citet{csillag2025prediction} adapt IPD ideas to the e-value framework, and building on conformal inference, \citet{candes2025probably} propose probably approximately correct label methods.

Lastly, while we focus on settings where the outcome is difficult or costly to measure, many IPD methods naturally extend beyond predicted outcomes to accommodate machine-predicted covariates, as well as blockwise and more general missingness patterns. These include previously discussed methods, such as PSPA~\citep{miao2023assumption}, PSPS~\citep{miao2024task}, and RePPI~\citep{ji2025predictions}, as well as more recent approaches, such as those by \citet{zhao2025imputation} and \citet{chen2025unified}. In addition, tailored methods such as those proposed by \citet{sondhi2023postprediction} and \citet{fong2021machine} focus specifically on noisy or misclassified surrogates for difficult-to-measure features. Across these disciplines, IPD methods share a common theme of combining a small set of gold-standard data with a large sample that has been augmented with predictions. Together, they showcase the versatility and growing importance of IPD in modern data-driven science.

\subsection{Our Case Studies, Revisited}

We now revisit our case studies in light of IPD. First, in the study of voter turnout patterns, we apply PPI~\citep{angelopoulos2023ppi++} to achieve IPD-corrected inference. To simulate a more realistic setting, we assume race was observed for only a 20\% subset of individuals. We then model $\Pr({\rm Race} = r \mid {\rm Voted} = 1)$ via logistic regression and apply PPI, treating the surname-based race probabilities as noisy surrogates for true race. We then use Bayes' Rule to obtain corrected estimates of $\Pr({\rm Voted} = 1 \mid {\rm Race} = r)$.

Table \ref{tab:turnout-stratified-ipd} shows that the PPI-corrected estimator reduces error across all groups relative to the naive approach, even when only 20\% of labels are observed. The table reports turnout probabilities under the true, na{\"i}ve, and IPD-corrected estimators, along with their absolute errors, improvement (IPD gain), and predictive metrics. In most groups, IPD reduces bias relative to the na{\"i}ve approach, notably for the White and Other race categories, where deviations from truth shrink by an order of magnitude. However, not all groups benefit equally: for example, the Black category exhibits a small increase in error after correction. This illustrates an important practical lesson: IPD methods introduce additional parameters to estimate, and their performance therefore depends on the size and representativeness of the labeled data and on the quality of the predictive model. While IPD can, on average, improve inferential validity, it does not guarantee uniform improvement in every subgroup or finite sample. Moreover, the magnitude of IPD's benefit appears largely independent of the predictive accuracy (e.g., AUC), reinforcing that high classification performance does not translate directly into better inference.

\begin{table}[!ht]
  \centering
  \caption{Turnout by race: true vs.~na{\"i}ve and IPD-corrected estimates. Errors are absolute deviations from truth; IPD gain = $|{\rm Naive~Err}|-|{\rm IPD~Err}|$ (positive = improvement). Higher Accuracy/AUC and larger IPD gain are better (blue = better, yellow = worse).}
  \label{tab:turnout-stratified-ipd}
  \vspace{0.5ex}
  \begin{tabularx}{\textwidth}{lrrrrrrrr}
    \toprule
    \textbf{Race} & \textbf{True} & \textbf{Naive} & \textbf{IPD} & \textbf{Naive Err} & \textbf{IPD Err} & \textbf{IPD Gain} & \textbf{Accuracy} & \textbf{AUC} \\
    \midrule
    White   & 0.7488 & 0.7392 & 0.7486 & \shadeabserr{-0.0096}{0.0175} & \shadeabserr{-0.0002}{0.0175} & \shadefromto{0.0093}{-0.0175}{0.0175} & \shadefromto{0.8013}{0.8016}{0.9853} & \shadefromto{0.8300}{0.5242}{0.9341} \\
    Black   & 0.7155 & 0.7145 & 0.7137 & \shadeabserr{-0.0010}{0.0175} & \shadeabserr{ -0.0018}{0.0175} & \shadefromto{-0.0008}{-0.0175}{0.0175} & \shadefromto{0.8811}{0.8016}{0.9853} & \shadefromto{0.8479}{0.5242}{0.9341} \\
    Hispanic& 0.6282 & 0.6336 & 0.6287 & \shadeabserr{ 0.0054}{0.0175} & \shadeabserr{ 0.0005}{0.0175} & \shadefromto{ 0.0050}{-0.0175}{0.0175} & \shadefromto{0.9422}{0.8016}{0.9853} & \shadefromto{0.9336}{0.5242}{0.9341} \\
    Asian   & 0.6100 & 0.6069 & 0.6087 & \shadeabserr{-0.0031}{0.0175} & \shadeabserr{-0.0013}{0.0175} & \shadefromto{ 0.0019}{-0.0175}{0.0175} & \shadefromto{0.9851}{0.8016}{0.9853} & \shadefromto{0.8547}{0.5242}{0.9341} \\
    Other   & 0.6384 & 0.6213 & 0.6327 & \shadeabserr{-0.0171}{0.0175} & \shadeabserr{-0.0058}{0.0175} & \shadefromto{ 0.0114}{-0.0175}{0.0175} & \shadefromto{0.9609}{0.8016}{0.9853} & \shadefromto{0.5244}{0.5242}{0.9341} \\
    \bottomrule
  \end{tabularx}
\end{table}

\begin{figure}[!ht]
    \centering
    \includegraphics[width=\linewidth]{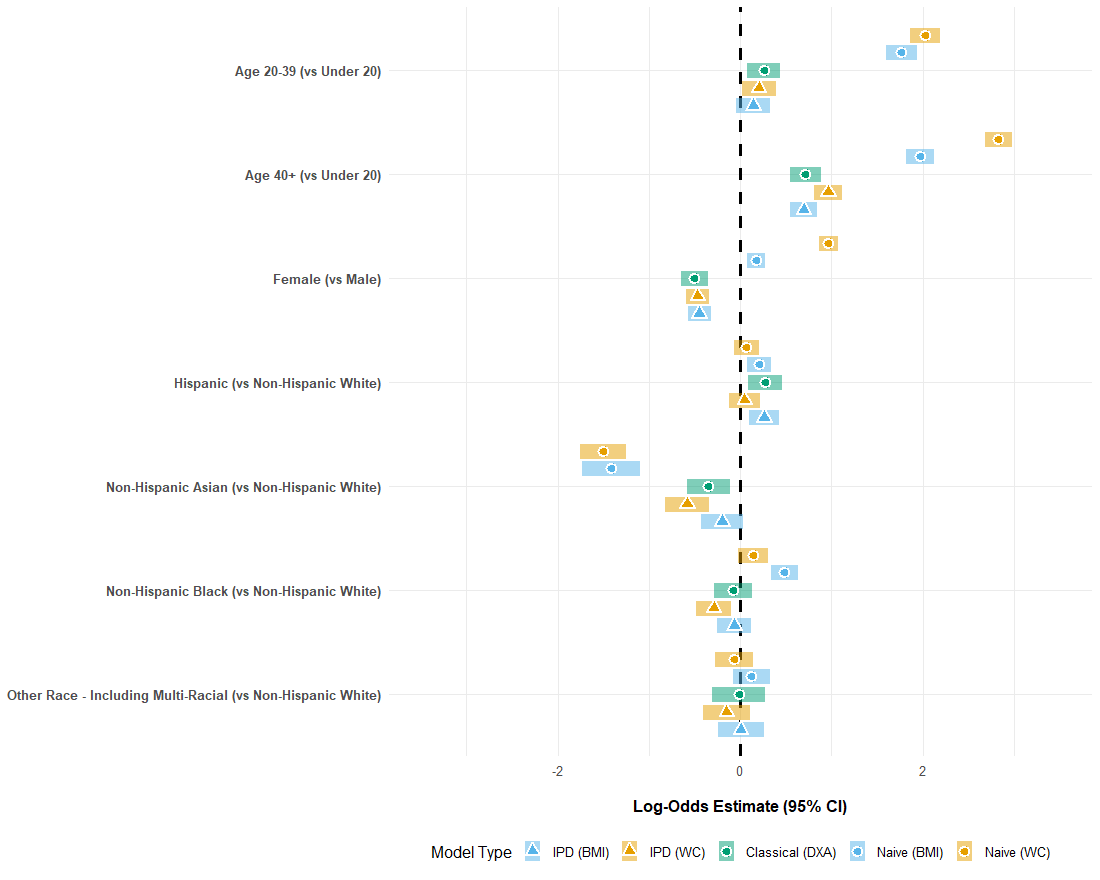}
    \caption{Coefficient (log-odds) estimates and 95\% confidence intervals for logistic regressions of obesity on certain demographic risk factors (age, sex, and race). Obesity is a binary outcome defined based on pre-specified thresholds for three continuous measures of adiposity: dual-energy X-ray absorptiometry (DXA)-based adiposity \% body fat (green), body mass index (BMI; kg/m$^2$, blue), and waist circumference (WC; cm, yellow), compared with and without IPD-based correction (triangular versus circular points).}
    \label{fig:bmi_forest2}
\end{figure}

Secondly, for our study on population-level inference for demographic risk factors of adiposity, we now apply PSPA~\citep{miao2023assumption} to both the naive logistic regression models which use BMI and WC, respectively, as proxies for DXA-based obesity. Here, we now combine the 2017-2018 and August 2021-August 2023 waves of NHANES, as the post-COVID wave did not measure DXA-based adiposity. As shown in Figure \ref{fig:bmi_forest2}, the IPD-corrected regression coefficients for BMI- and WC-based obesity align more closely with the DXA-based model that uses only the 2017-2018 data, as expected.

Here, we see that BMI and WC are convenient, but imperfect proxies for DXA-measured percent body fat. The naive regressions that treat BMI- or WC-based obesity as the true outcome introduce bias into estimates of how obesity relates to risk factors like age, sex, and race/ethnicity. In contrast, classical regression on the gold‐standard DXA measurements yields unbiased associations, but at the expense of substantial cost and limited sample size. IPD bridges this gap: by leveraging DXA-based percent body fat in a small labeled cohort, and then applying IPD corrections to a much larger BMI/WC-only cohort, one can recover unbiased effect estimates with potentially greater precision than the classical approach. More broadly, understanding how different surrogate measures influence population-level inference is critical for designing robust epidemiologic and clinical studies in which direct measurement of complex health outcomes may be infeasible.

Together, these results reinforce our message: while predictions may be useful, treating them as objective truth in downstream inference can yield misleading conclusions. The IPD framework offers principled methods for recovering valid estimates, even with limited gold-standard data.


\section{Doesn't This Sound Familiar?}
\label{sec:4}

{\it `The author [...] studied some types of [...] technique which consists of obtaining an expression for a character $y$, sometimes difficult or uneconomic to measure directly, in terms of an appreciable correlated character $x$, easier to obtain.'} \hfill $\sim$\ \citet{bose1943note}

\vspace{2ex}

Bose wrote this in reference to double sampling theory from nearly a century ago. If this sounds similar to the setup we have just introduced for conducting inference with predicted data, it should. Although IPD has only recently emerged as a distinct subfield, its core ideas have deep roots in survey sampling, missing data, causal inference, semi-supervised learning, econometrics, and measurement error modeling (see Figure~\ref{fig:comparison} for some select fields and Figure~\ref{fig:compare_methods_full} in Appendix~\ref{sec:B}). What distinguishes IPD in the modern era is not the idea of using auxiliary information, but the reality that auxiliary information now often comes from pre-trained, black-box algorithms.

In practice, researchers typically lack access to the algorithm's training data, whether because of proprietary restrictions (e.g., large language models), privacy concerns (e.g., medical records), or simple infeasibility. Without knowledge of the algorithm's training set or internal uncertainty estimates, users cannot reliably quantify the prediction error in advance or design studies around it. Equally important, the end-users do not know the model's architecture or its operating characteristics. For both modern predictive algorithms (e.g., deep neural networks, large language models, ensembles) and simple prediction rules (e.g., our BMI example), the linear or semi-parametric assumptions behind standard auxiliary or imputation strategies are incorrect or unverifiable. Consequently, IPD methods all share an `assumption-lean' and `data-adaptive' paradigm~\citep{miao2023assumption}. This model-agnostic approach to bias and variance adjustment enables methods for IPD to deliver valid, efficient inference with minimal assumptions.

\begin{figure}
    \centering
    \includegraphics[width=\linewidth]{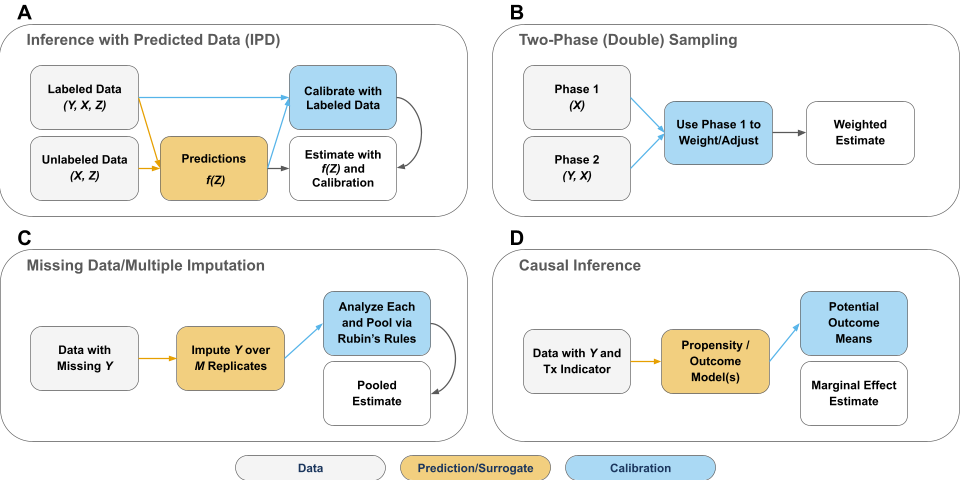}
    \caption{Comparison of the workflow for inference with predicted data (IPD; Panel A) to several foundational fields in statistics, including two-phase sampling (Panel B), missing data/multiple imputation (Panel C), and causal inference (Panel D). }
    \label{fig:comparison}
\end{figure}

As seen, the connection to classical two-phase (double) sampling is especially apparent. In double sampling, the investigator measures inexpensive auxiliary variables in a large, first-phase sample and then collect the costly outcome measure in a smaller, second-phase sample to achieve higher precision at lower costs~\citep{neyman1938contribution, cochran1939use, bose1943note, srivastava2016historical, kubiak2022prior}. IPD differs in that our auxiliary variable is the machine predictions, $\hat{Y}$, but these concepts still hold. Double sampling theory shows that measuring auxiliary data in a large first-phase sample and the costly outcome in a smaller second-phase subsample can yield more precise estimates at a reduced cost via optimal allocation. 

Recent works extend these ideas further to active sampling. \citet{angelopoulos2025cost} formulate model evaluation as a budget-constrained sampling problem to select which unlabeled points to verify. Similarly, \citet{zrnic2024active} use a black-box predictor to guide the selection of data points whose labels would most reduce inferential uncertainty. \citet{eyre2024auto} develop an approach for the continuous evaluation of large generative model using robust regression in few-label regimes. This modern take on guided sampling extends the classic two-phase design into active IPD, where labeling decisions are AI/ML-driven and optimized for downstream inference.

The same logic underpins the missing data literature, as we `impute' $Y$ using auxiliary information from the unlabeled $\boldsymbol{Z}$. Naively replacing $Y$ with one realization of $\hat{Y} = \hat{f}(\boldsymbol{Z})$ amounts to `single imputation,' which is known to underestimate uncertainty~\citep{little2019statistical}. IPD instead treats $\hat{Y}$ as a noisy surrogate~\citep{ji2025predictions} and propagates its uncertainty into downstream inference. PostPI estimates this variability using a `relationship' model, while PPI-based methods directly regress the residual difference, $\hat{Y} - Y$, on $\boldsymbol{X}$ to obtain a `rectified' variance estimate. PSPA, PDC, Chen \& Chen, and others operate similarly. IPD methods have analogs to multiple imputation as well, where pooling results from independent draws of $\hat{Y}$ captures both within- and between-imputation variability ~\citep{rubin1976inference}. \citet{wang2020methods} propose a bootstrap-based PostPI correction that averages results over multiple generated pseudo-outcomes. Similarly, \citet{zrnic2024note} uses bootstrap resampling of the data to create multiple debiased estimators, correcting on the parameter space, rather than the outcome space. More recently,~\citep{miao2024task} introduce a task-agnostic framework which bootstraps summary statistics to learn the covariance between the true and predicted outcomes for debiasing and variance inflation.

IPD has a natural analog in causal inference, which treats unobserved counterfactuals as `missing' outcomes. Classical causal estimators, such as inverse probability weighting (IPW) and augmented IPW (AIPW), leverage models for exposure status (here, the `propensity' for being labeled) and/or the outcome to recover unbiased effect estimates~\citep{rubin2005causal}. Analogously, many IPD estimators assume the outcome is MCAR, where every unit has the same probability, or {\it propensity}, of being labeled, $\pi = \tfrac{1}{1 + r} = \tfrac{1}{1 + n_l/n_u}$. Under this view, PPI-like estimators are special cases of IPW estimators with either/both a constant propensity model, $\pi$, and a working outcome model, $\hat{f}(\boldsymbol{Z})$, which they use to form debiased estimators. This perspective clarifies both why IPD can achieve unbiasedness, and how MAR extensions can arise by letting $\pi(\boldsymbol{X})$ vary with covariates.

More specifically, this type of correction relates to AIPW estimators, which are {\it doubly robust} (i.e., consistent if either the propensity or outcome model is correct) and {\it semiparametrically efficient} when both models are well-estimated~\citep{robins1994estimation}. AIPW estimators arise from the efficient influence function for a target parameter, guiding the construction of one-step and targeted-minimum-loss estimators that achieve the smallest possible asymptotic variance. \citet{ji2025predictions} show that many IPD estimators are solutions to a modified AIPW estimating equation,

\begin{equation}
\label{eq:aipw}
    \sum_{i=1}^{n_l+n_u}\left\{\frac{D_i}{\pi} U_\theta\left(\boldsymbol{X}_i, Y_i\right)-\frac{D_i-\pi}{\pi} \psi_\theta\left[\boldsymbol{X}_i, \hat{f}(\boldsymbol{Z_i)}\right]\right\} = \boldsymbol{0},
\end{equation}
where $U_\theta(\cdot)$ is a score function, $\psi_\theta(\cdot)$ is user-specified, $Y$ is missing with probability $\pi$, and $D_i \in\{0,1\}$ indicates whether $Y$ is observed. This is essentially \eqref{eq:unifying} with convex $\ell_\theta(\cdot)$ and $\psi_\theta(\cdot) = \nabla g_\theta(\cdot)$. Intuitively, one component models the labeling mechanism (propensity), and the other models the relationship between $\hat{f}(\boldsymbol{Z})$ and $Y$. \eqref{eq:aipw} also gives intuition for when IPD is useful. The optimal $\psi_\theta(\cdot)$ is $\psi_\theta^\star[\boldsymbol{X}_i, \hat{f}(\boldsymbol{Z}_i)] = \mathbb{E}[U_\theta(\boldsymbol{X}, Y) \mid \boldsymbol{X}, \hat{f}(\boldsymbol{Z})]$, which shows that na{\"i}vely substituting $\hat{f}(\boldsymbol{Z})$ for $Y$ does not yield a valid estimator, except in the extreme case where $\hat{f}(\boldsymbol{Z}) = \mathbb{E}[Y \mid \boldsymbol{X}]$. Moreover, the solution to \eqref{eq:aipw} is semiparametrically efficient only if $\psi_\theta(\cdot) = \psi_\theta^\star(\cdot)$. This is why most IPD are only efficient among the class of `single imputation-based' estimators~\citep{angelopoulosnote, ji2025predictions, xu2025unified}, while others~\citep{gronsbell2024another, ji2025predictions} are semiparametrically efficient if $\phi_\theta(\cdot)$ is consistently estimated, but is always at least as efficient as the classical approach.

IPD naturally connects with the surrogate outcomes literature as well. \citet{ji2025predictions} formalize this in the AI era by showing that many of the IPD methods emerge as special cases of a general influence function-based estimator \eqref{eq:unifying}, whose augmentation term corrects for surrogate bias and variance. RePPI further learns an optimal `imputed loss' linking $\hat{f}(\boldsymbol{Z})$ and $Y$, achieving semiparametric efficiency while maintaining valid inference. The same logic underlies measurement error models, which treat observed proxies as noisy versions of the true latent variables and correct via regression calibration or other techniques~\citep{chen2005measurement, stefanski2000measurement, fuller2009measurement}.

IPD is also a special form of semi-supervised learning (SSL), which takes a small {\it labeled} set and augments it with a larger {\it unlabeled} set whose missing outcomes are filled in by model predictions~\citep{van2020survey}. Cross-PPI~\citep{zrnic2024cross} bridges the gap between IPD and SSL and employs cross-fitting~\citep{chernozhukov2018double} to mitigate overfitting bias in small labeled samples. In this spirit, RePPI~\citep{ji2025predictions} and DSL~\citep{egami2023using} also extend IPD by generating bias-corrected pseudo-outcomes via cross-fitting a supervised machine learning model for $Y$ on $\hat{f}(\boldsymbol{Z})$, $\boldsymbol{Z}$, and $\boldsymbol{X}$. They then apply a doubly-robust estimator that combines the pseudo-outcomes with the limited gold-standard labels and guarantees asymptotic unbiasedness, even when their supervised machine learning model is misspecified. 

More broadly, \citet{xu2025unified} present a unified semiparametric framework for semi-supervised learning that both characterizes when unlabeled data can improve inference and constructs two classes of estimators: a `safe' estimator that never underperforms the supervised baseline, and an `efficient' estimator that, under stronger assumptions, attains the semiparametric efficiency bound. Their theory subsumes IPD methods like PPI, PPI++, PSPA, PDC, and Chen and Chen by showing that, for arbitrary inferential targets, one can explicitly quantify the efficiency gain from predictions and design influence function-based one-step estimators that optimally leverage both labeled and unlabeled data. Together, these works demonstrate how principled sampling designs and semiparametric efficiency theory can guide both the construction and the theoretical assessment of IPD methods that bridge machine learning predictions and valid statistical inference.

In other areas, economists have long employed the idea of using a simpler `auxiliary' model to stand in for an intractable structural model. In {\it indirect inference}~\citep{gourieroux1993indirect, smith2008indirect, efron2010future}, one fits an auxiliary regression to both the real data and data simulated under candidate parameters, then chooses the parameters that make the auxiliary model estimates agree. This matching step corrects for the auxiliary model's bias, in a similar spirit to IPD's calibration of predicted outcomes. This underscores how surrogate models have concurrent foundations in many econometric applications~\citep{le2016testing, bruins2018generalized}. Likewise, {\it data-augmented regression}~\citep[DAR;][]{hooker2012prediction}, constructs new features from cross-validated predictions and includes them with the original covariates in a regularized regression. These predicted features act as penalty, as opposed to penalizing for sparsity, to stabilize the downstream regression. DAR's use of out-of-sample predictions to enrich a parametric model anticipates IPD's strategy of treating predictions as surrogates and adjusting for their uncertainty and bias in the inferential stage.


\section{Some Limitations and Open Problems}

Despite rapid progress, IPD still faces important limitations and open questions. Current methods are promising and theoretically grounded, but do not yet cover all practical scenarios.

{\it Optimal Efficiency.} Recent works have studied semiparametric efficiency limits for IPD~\citep{angelopoulos2024note, gronsbell2024another, ji2025predictions, xu2025unified}. These results clarify when predictions can improve precision and when they cannot, and show that even optimal IPD estimators do not uniformly dominate classical semi-supervised procedures. These works outline a unifying framework for IPD and explore the conditions under which using AI/ML-generated data can confer any efficiency gain over traditional supervised inference. Understanding this is an important area of ongoing research. Moreover, \citet{mani2025no} show that the asymptotic behavior of these methods does not hold in finite samples, reinforcing that there is `no free lunch' in practice.

{\it Generalizability, Transportability, and Distribution Shift.} Most IPD procedures assume the outcome is MCAR, or that the labeled subset is representative of the unlabeled population. In practice, label selection can depend on observed (MAR) or unobserved features, and predictors may behave differently across subgroups. Some methods already relax the MCAR assumption~\citep{mccaw2024synthetic, gronsbell2024another, fisch2024stratified, egami2023using, chen2025unified}, but robustness to covariate shift and selection bias is still developing. Transfer learning and covariate balancing ideas may extend IPD when the labeled set is not a random subsample.

{\it Prospective AI/ML-Enriched Designs.} Another promising direction involves designing prospective studies that explicitly leverage the availability of, or intent to use, predictive models to enrich limited data sources. Power and sample size calculations that incorporate predicted quantities and their uncertainty could improve study efficiency and reduce labeling costs. Prior work suggests that improving surrogate fidelity can outweigh simply increasing $n$, especially with biased non-probability data~\citep{miao2024valid, mccaw2024synthetic, gronsbell2024another}. As modern studies blend probability samples, convenience samples, and synthetic data~\citep{raghunathan2021synthetic, pezoulas2024synthetic, chen2025generating}, IPD offers a lens for calibrating mixed data sources.

{\it Inference for Complex Parameters.} Most IPD methods target means, quantiles, or regression parameters. Extending to survival, longitudinal, spatial, time-series, and nonparametric models is ongoing. In clinical trials, causal IPD perspectives~\citep{mukherjee2024causal, gradu2025valid, demirel2024prediction} and surrogate outcome formulations~\citep{ji2025predictions} are promising. Multiple testing control~\citep{mccaw2024synthetic, miao2024valid} for IPD is another important area.

{\it Limited or No Ground Truth Data.} IPD relies on at least a small labeled set for calibration. In unsupervised or latent class settings, labels may be unavailable, precluding rigorous inference without additional assumptions. Approaches such as PSPS~\citep{miao2024task} or federated PPI~\citep{luo2024federated} can operate with restricted summaries, but robust inference with no labels remains challenging. In tasks like 3D protein prediction or single-cell clustering, uncertainty stems from structural/modeling assumptions rather than noisy labels. Clarifying how IPD interfaces with these settings, and how it differs from post-selection inference, poses conceptual and practical questions. This is a unique case where two perspectives share a common space in application, but theoretically target two different statistical quantities, a topic we will explore in future work. 

{\it High-Dimensional Data.} Another point of common ground between post-selection inference and IPD may be in high-dimensional data problems. Most IPD theory assumes low-dimensional downstream models so that classical inference is feasible. When the analysis itself is high-dimensional, new issues arise: combining IPD with sparsity tools (e.g., LASSO) and delivering valid post-selection inference when predictions guide model choice remains largely unexplored. Bridging IPD with post-selection frameworks is another potentially interesting avenue for future work.

{\it Communication and Trust.} Lastly, IPD adds complexity to analysis pipelines, potentially hindering uptake from stakeholders and researchers from diverse backgrounds. Concepts like accurate predictions not conferring additional information or the rationale behind corrected confidence intervals being potentially wider or narrower can be unintuitive, even to experts. Developing accessible educational materials, clear diagnostics, and intuitive visualizations to confirm that IPD adjustments function correctly is crucial. To this end, to promote open and collaborative science, we have developed an R package implementing many of the methods referenced throughout~\citep{salerno2024ipd}. This package (1) provides easy access for researchers wanting to apply these methods in practice, and (2) enables data scientists working in this area to develop the field and compare their approaches to current methods. However, increased adoption in applied settings will depend on our ability to demonstrate the reliability, interpretability, and practical utility of IPD methods. Addressing these remaining challenges, and more, will define the future trajectory of IPD. 


\section{Final Thoughts and Conclusions}

As AI/ML continues to reshape many scientific domains, the challenge of statistical inference requires immediate and thoughtful attention. This review is meant to shed light on both the pitfalls of treating AI/ML-generated data as observed and the promise of recent methods for IPD. We briefly highlighted core methods, discussed subsequent enhancements, and noted emerging unifying theory. These developments empower researchers across disciplines to rigorously incorporate AI/ML-generated predictions into statistical analyses without compromising inferential validity.

{\it Veridical Data Science} offers a principled lens to view IPD, by emphasizing {\it predictability}, {\it computability}, and {\it stability} (PCS) as essential components of trustworthy scientific practice~\citep{yu2020veridical, yu2024veridical}. For IPD, {\it predictability} concerns both the upstream prediction model and downstream statistical model's ability to yield valid and replicable conclusions. As we saw, high predictive accuracy alone can be misleading when it does not preserve the estimand of interest. {\it Computability}, ensures that analytic pipelines can be executed at scale, something that is increasingly relevant as predictive models grow in size and complexity, and as a greater volume of data can be imputed or synthetically generated. IPD allows for valid statistical inference with only a modest labeled dataset by compressing large training data into compact surrogates and correcting for bias and uncertainty in a statistically principled way. {\it Stability} concerns the robustness of findings under reasonable data perturbations or modeling choices. Although many IPD methods do not make assumptions about the internal workings, quality, or training data of the upstream predictive model, they attempt to measure {\it stability} via data-adaptive mechanisms. These mechanisms capture how sensitive inference is to prediction error and inflate standard error estimates to reflect this additional uncertainty. In this way, IPD provides model agnostic {\it stability}.

{\it Stability}, however, is broader than valid surrogacy. Design and data collection matter. We must ask questions such as `are our the labeled data representative?,' `do selection mechanisms bias both prediction and inference?,' and `are our conclusions generalizable, or do models trained in one cohort fail elsewhere?' Fairness and algorithmic bias also warrant explicit auditing within the IPD framework~\citep{xu2024addressing, gao2024semi, gao2025reliable}.  Further, modeling choices beyond the `given' prediction function, such as the form of the downstream regression or the method of variance estimation, can also influence the results of the overall analysis. These choices also extend to practical considerations, such as label costs and privacy requirements in data sharing, which can limit the extent to which we can study {\it stability}. While IPD seeks to address one aspect of {\it stability}, trustworthy scientific practice demands a holistic approach to the PCS framework.

Lastly, veridical practice also requires {\it transparent} reporting of modeling choices, thoughtful consideration of algorithmic bias, and systematic checks to ensure that scientific findings derived from predicted data are not only accurate, but valid, interpretable, reproducible, and meaningful. As we have tried to illustrate, these principles are foundational to emerging IPD methods and will remain central as AI/ML tools become increasingly embedded in empirical research. For IPD in particular, transparency includes what was labeled, when and how units were selected for labeling, and the assumptions linking gold-standard measurements to predictions. 

At the outset, we asked two questions: `do we really even need data?,' and `how would you draw a rhinoceros if you had never seen one?' Our question of `how would you draw a rhino?' underscores a central theme: like the artists, today's data scientists must navigate uncertainty, bias, and the limits of their tools. The critical question is not whether we can generate accurate predictions, but whether these predictions carry the correct information for the intended scientific inquiry. Our titular question, `do we really even need data?', is intentionally provocative, but not rhetorical. Black-box predictions cannot replace carefully measured data from well-designed studies. Yet, used thoughtfully, predictions can extend scarce gold-standard labels, improve efficiency, and broaden the scope of research. Many of the principles we emphasize trace back to classical statistics, and adapting them to today's AI-driven landscape requires careful experimental design, valid inferential techniques, rigorous evaluation, and open, interdisciplinary practice.


\subsection*{Disclosure Statement}

JTL reports Coursera courses which generate revenue for both Johns Hopkins University and the Fred Hutchinson Cancer Center. JTL reports co-founding and serving on the board of Synthesize Bio.


\subsection*{Acknowledgments}

We thank C.M.~K{\"o}semen for the graciously allowing us to reprint his artwork.

 
\subsection*{Contributions}

SS: Conceptualization, Methodology, Formal Analysis, Investigation, Writing - Original Draft, Writing - Review \& Editing, Visualization; KH: Conceptualization, Methodology, Formal Analysis, Investigation, Writing - Original Draft, Writing - Review \& Editing, Visualization; AA: Methodology, Investigation, Writing - Review \& Editing; AN: Methodology, Investigation, Writing - Review \& Editing; THM: Supervision, Conceptualization, Methodology, Project Administration, Writing - Review \& Editing, Funding Acquisition; JTL: Supervision, Conceptualization, Methodology, Project Administration, Writing - Review \& Editing, Funding Acquisition.


\subsection*{Generative AI Disclosure Statement}

We utilized Generative AI (OpenAI's GPT-4/5) in the production of this manuscript, in the following ways: (1) proposing sentences to include in the manuscript and (2) iteratively improving the concision and clarity of the writing. We have carefully reviewed all aspects of the manuscript for accuracy and coherence. All scientific insights, analysis
and interpretation of data and scientific conclusions are made solely by the authors. All errors are our own. This disclosure is adapted from~\citet{visokay2025measure}.

\newpage


\appendix


\section{Additional Analytic Results}
\label{sec:A}

\subsection{Additional Derivations for the Illustrative Example}

\subsubsection{Setup and Notation}

Recall that we have predictors, $\boldsymbol{Z} = (Z_1, \ldots, Z_{10})$ with $Z_j \stackrel{{\rm iid}}{\sim} \mathcal{N}(0, 1)$. We define our outcome as $Y = \sum_{j=1}^{10} Z_j + \varepsilon$, with $\varepsilon \sim \mathcal{N}(0,1)$ independent of $\boldsymbol{Z}$. We form predicted outcomes, $\hat{Y}^{(m)} = \hat{f}^{(m)}(\boldsymbol{Z})$, using a fixed prediction rule from a pre-trained model (e.g., a random forest) trained on an independent sample, where $m$ is the set of features used by a given prediction rule. Our downstream inferential model is the simple linear regression of $Y$ on $Z_1$, 
\[
    Y = \beta_0 + \beta_1 Z_1 + \epsilon,
\]
and our target is the slope parameter, $\beta_1$ for $Z_1$. We denote the target of the na{\"i}ve regression of $\hat{Y}^{(m)}$ on $Z_1$ by $\eta_1$, whereas the target for the true data is the marginal slope of $Y$ on $Z_1$, which is
\[
    \beta_1 = \frac{{\rm Cov}(Y, Z_1)}{{\rm Var}(Z_1)} = 1.
\]

\subsubsection{Analytic Bias}

We write the $L_2$ projection of $\hat{f}^{(m)}(\boldsymbol{Z})$ onto ${\rm span}({1, Z_1})$ as
\[
    \hat{f}^{(m)}(\boldsymbol{Z}) = a_0 + a_1 Z_1 + g(\boldsymbol{Z}_{-1}); \quad \mathbb{E}[Z_1 g(\boldsymbol{Z}_{-1})] = 0,
\]
where $\boldsymbol{Z}_{-1} = (Z_2, \ldots, Z_{10})$ and $g(\cdot)$ collects the non-$Z_1$ part (including all nonlinearity and dependence on $\boldsymbol{Z}_{-1}$). Then, the na{\"i}ve slope is given by
\[
    \eta^{(m)}_1 = \frac{{\rm Cov}[Z_1, \hat{f}^{(m)}(\boldsymbol{Z})]}{{\rm Var}(Z_1)} = \mathbb{E}[Z_1 \hat{f}^{(m)}(\boldsymbol{Z})] = a_1 + \underbrace{\mathbb{E}[Z_1 g(\boldsymbol{Z}_{-1})]}_{= 0 \text{ by construction}} = a_1.
\]
Thus, the bias, in general, is 
\[
    {\rm Bias}(\eta_1) = a_1 - 1.
\]
For a random forest trained on a subset of features, $S \subseteq{1, \ldots, 10}$, is (population-wise) consistent for $m_S(\boldsymbol{Z}) := \mathbb{E}[Y \mid \boldsymbol{Z}_S]=\sum_{j\in S} Z_j$ under standard regularity conditions and with and sufficient data. With finite samples and regularization, random forests typically exhibit attenuation, so
\[
    \hat{f}(\boldsymbol{Z}) \approx c_S \sum_{j\in S} Z_j + r_S(\boldsymbol{Z}); \quad c_S \in(0,1]; \quad \mathbb{E}[\boldsymbol{Z}, r_S(\boldsymbol{Z})] = 0,
\]
where $c_S$ captures shrinkage and $r_S$ is an orthogonal remainder that is typically small for well-trained random forests. Since the $Z_j$ are independent and mean-zero,
\[
  \eta_1 = \mathbb{E}[Z_1 \hat{f}(\boldsymbol{Z})] \approx c_S \mathbb{E}\left[Z_1 \sum_{j\in S} Z_j\right] = \begin{cases} c_S, & 1\in S,\\ 0, & 1 \notin S, \end{cases}
\]
so
\[
  {\rm Bias}(\eta_1) \approx \begin{cases} c_S - 1, & 1\in S,\\ -1, & 1\notin S. \end{cases}
\]

\subsubsection{Residual Variance}

The residual variance of the na{\"i}ve regression is
\[
    \sigma_\eta^2  = {\rm Var}[\hat{f}(\boldsymbol{Z}) -a_0 - a_1 \boldsymbol{Z}_1] = {\rm Var}[g(\boldsymbol{Z}_{-1})].
\]
So the na{\"i}ve standard error for $\eta_1$ (ignoring finite-sample degrees of freedom) behaves like
\[
    SE(\eta_1) \asymp \sqrt{\frac{\sigma_\eta^2}{n {\rm Var}(Z_1)}} = \sqrt{\frac{{\rm Var}[g(\boldsymbol{Z}_{-1})]}{n}},
\]
where the oracle regression of $Y$ on $Z_1$ has residual variance
\[
    \sigma^2 = {\rm Var}\left(\sum_{j \neq 1}\beta_j Z_j + \varepsilon\right) = \sum_{j\neq 1} \beta_j^2 + \sigma_\varepsilon^2 = (p - 1) + 1 = p,
\]
in our example. Its standard error scales as $\sqrt{p/n}$. Typically, ${\rm Var}[g(\boldsymbol{Z}_{-1})] < p$, so the na{\"i}ve inference will underestimate the standard error, even when $\eta_1$ is close to the truth. Using random forests, the projection residual is approximately
\[
    g(\boldsymbol{Z}_{-1}) \approx c_S \sum_{j\in S\setminus{1}} Z_j + r_S(\boldsymbol{Z}),
\]
so
\[
    \sigma_\eta^2 = {\rm Var}[g(\boldsymbol{Z}_{-1})] \approx c_S^2 |S\setminus{1}| + {\rm Var}[r_S(\boldsymbol{Z})].
\]
Compared to the oracle residual variance, $p$, when $|S| < p$, $\sigma_\eta^2$ is smaller than $p$, producing anticonservative standard errors. When $\hat{f}$ includes $Z_1$ but omits many other predictors, $\eta_1 \approx c_S < 1$ (attenuation bias), with $\sigma_\eta^2 \approx c_S^2(|S| - 1)$ (underestimated variance). Excluding $Z_1$ entirely gives $\eta_1 \approx 0$ (strong bias) and $\sigma_\eta^2 \approx c_S^2|S|$. So the inferential behavior of the na{\"i}ve regression is governed by two projection objects:
\[
    a_1 = \arg\min_{a}\mathbb{E}\{[\hat{f}(\boldsymbol{Z}) - a Z_1 - b]^2\},
\]
and
\[
    {\rm Var}[g(\boldsymbol{Z}_{-1})] = \min_{a,b}\mathbb{E}\{[\hat{f}(\boldsymbol{Z}) - a Z_1 - b]^2\}.
\]

\subsubsection{Finite Sample Leakage}

As another brief point of consideration, random forests based on finite samples can induce weak correlation between $Z_1$ and the part of $\hat{f}$ built from $\boldsymbol{Z}_{-1}$. Let
\[
    \hat{f}(\boldsymbol{Z}) = a_0 + a_1 Z_1 + g(\boldsymbol{Z}_{-1}); \quad \ell := \mathbb{E}[Z_1 g(\boldsymbol{Z}_{-1})].
\]
Then, the na{\"i}ve slope is $\eta_1 = a_1 + \ell$, and the residual variance is
\[
    \sigma_\eta^2 = {\rm Var}(\hat{f}) - \eta_1^2 = {\rm Var}(a_1 Z_1 + g) - (a_1 + \ell)^2 = {\rm Var}(g) -\ell^2,
\]
so any leakage, $|\ell| > 0$ both biases the slope estimate and further reduces the residual variance. Cauchy-Schwarz bounds $|\ell| \leq \sqrt{{\rm Var}(g)}$, but in practice $|\ell|$ is typically small. Empirically, $\ell$ can be estimated on $\mathcal{L}$ by regressing $\hat{f}(\boldsymbol{Z})$ on $(1, Z_1)$ and inspecting the orthogonality diagnostics.

\newpage


\section{Additional Comparisons}
\label{sec:B}

\begin{figure}[!ht]
    \centering
    \includegraphics[width=0.8\linewidth]{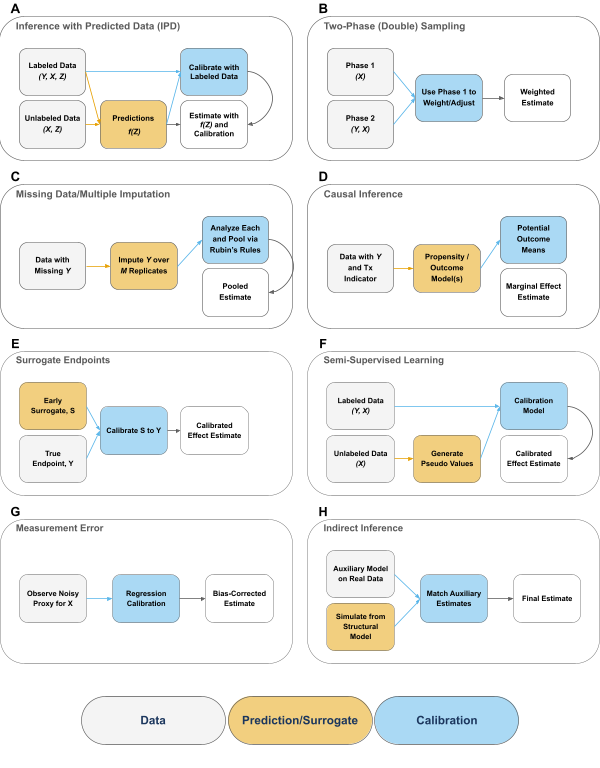}
    \caption{Comparison of the workflow for inference with predicted data (IPD; Panel A) to several foundational fields in statistics, including two-phase sampling (B), missing data/multiple imputation (C), causal inference (D), surrogate endpoints (E), semi-supervised learning (F), measurement error (G), and indirect inference (H).}
    \label{fig:compare_methods_full}
\end{figure}

\newpage


\bibliographystyle{abbrvnat}
\bibliography{references}


\end{document}